\documentclass[Afour,sageh,times]{sagej}
\usepackage{moreverb,url}
\usepackage[colorlinks,bookmarksopen,bookmarksnumbered,citecolor=red,urlcolor=red]{hyperref}
\newcommand\BibTeX{{\rmfamily B\kern-.05em \textsc{i\kern-.025em b}\kern-.08em
T\kern-.1667em\lower.7ex\hbox{E}\kern-.125emX}}
\def\volumeyear{2023}
\usepackage{bm}
\usepackage{amssymb}
\usepackage{amsthm}
\usepackage{mathtools}
\usepackage{nicefrac}  

\usepackage[noend]{algpseudocode}
\usepackage{algorithm}
\usepackage{setspace}

\usepackage{booktabs}
\usepackage{multicol}
\usepackage{multirow}
\usepackage{tabularx}
\usepackage{etoolbox}
\newrobustcmd{\B}{\bfseries}  
\usepackage{siunitx}  
\usepackage{diagbox}
\usepackage{pifont}  
\usepackage{todo}  
\usepackage[table]{xcolor}
\definecolor{red}{RGB}{200,50,50}
\definecolor{green}{RGB}{50,200,50}
\definecolor{blue}{RGB}{50,50,200}
\definecolor{orange}{RGB}{243,147,82}
\definecolor{purple}{RGB}{150,100,250}
\definecolor{yellow}{RGB}{231,198,100}
\definecolor{gray}{RGB}{98,95,75}
\definecolor{white}{RGB}{255,255,255}
\usepackage{lipsum}
\usepackage{times}

\makeatletter
\newcommand*{\T}{{\mathpalette\@transpose{}} }
\newcommand*{\@transpose}[2]{\raisebox{\depth}{$\m@th#1\intercal$}}
\makeatother
\DeclarePairedDelimiterX{\norm}[1]{\lVert}{\rVert_{2}}{#1}
\DeclareMathOperator*{\argmax}{arg\,max}
\DeclareMathOperator*{\argmin}{arg\,min}

\newcommand{\real}{\mathbb{R}}
\makeatletter
\newcommand\thefontsize{The current font size is: \f@size pt}
\makeatother

\usepackage{cleveref}  
\theoremstyle{definition}
\newtheorem{problem}{Problem}
\theoremstyle{definition}

\theoremstyle{definition}
\newtheorem{definition}{Definition}
\theoremstyle{definition}

\theoremstyle{definition}

\newtheorem{proposition}{Proposition}
\crefname{proposition}{Proposition}{Propositions}
\usepackage{balance}  
\apptocmd{\sloppy}{\hbadness 10000\relax}{}{}

\usepackage{graphicx}
\usepackage[caption=false,font=footnotesize]{subfig}
\usepackage{wrapfig}  
\usepackage{tikz}
\usetikzlibrary{positioning,shapes,bayesnet,calc,backgrounds}
\newcommand*\circled[1]{\tikz[baseline=(char.base)]{\node[shape=circle,draw,inner sep=2pt] (char) {#1};}}
\makeatletter
\newif\if@showgrid@grid
\newif\if@showgrid@left
\newif\if@showgrid@right
\newif\if@showgrid@below
\newif\if@showgrid@above
\tikzset{%
  every show grid/.style={},
  show grid/.style={execute at end picture={\@showgrid{grid=true,#1}}},%
  show grid/.default={true},
  show grid/.cd,
  labels/.style={font={\sffamily\small},help lines},
  xlabels/.style={},
  ylabels/.style={},
  keep bb/.code={\useasboundingbox (current bounding box.south west) rectangle (current bounding box.north west);},
  true/.style={left,below},
  false/.style={left=false,right=false,above=false,below=false,grid=false},
  none/.style={left=false,right=false,above=false,below=false},
  all/.style={left=true,right=true,above=true,below=true},
  grid/.is if=@showgrid@grid,
  left/.is if=@showgrid@left,
  right/.is if=@showgrid@right,
  below/.is if=@showgrid@below,
  above/.is if=@showgrid@above,
  false,
}
\def\@showgrid#1{%
  \begin{scope}[every show grid,show grid/.cd,#1]
    \if@showgrid@grid
      \begin{pgfonlayer}{background}
        \draw [help lines]
        (current bounding box.south west) grid
        (current bounding box.north east);
        \pgfpointxy{1}{1}%
        \edef\xs{\the\pgf@x}%
        \edef\ys{\the\pgf@y}%
        \pgfpointanchor{current bounding box}{south west}
        \edef\xa{\the\pgf@x}%
        \edef\ya{\the\pgf@y}%
        \pgfpointanchor{current bounding box}{north east}
        \edef\xb{\the\pgf@x}%
        \edef\yb{\the\pgf@y}%
        \pgfmathtruncatemacro\xbeg{ceil(\xa/\xs)}
        \pgfmathtruncatemacro\xend{floor(\xb/\xs)}
        \if@showgrid@below
          \foreach \X in {\xbeg,...,\xend} {
            \node [below,show grid/labels,show grid/xlabels] at (\X,\ya) {\X};
          }
        \fi
        \if@showgrid@above
          \foreach \X in {\xbeg,...,\xend} {
            \node [above,show grid/labels,show grid/xlabels] at (\X,\yb) {\X};
          }
        \fi
        \pgfmathtruncatemacro\ybeg{ceil(\ya/\ys)}
        \pgfmathtruncatemacro\yend{floor(\yb/\ys)}
        \if@showgrid@left
          \foreach \Y in {\ybeg,...,\yend} {
            \node [left,show grid/labels,show grid/ylabels] at (\xa,\Y) {\Y};
          }
        \fi
        \if@showgrid@right
          \foreach \Y in {\ybeg,...,\yend} {
            \node [right,show grid/labels,show grid/ylabels] at (\xb,\Y) {\Y};
          }
        \fi
      \end{pgfonlayer}
      \fi
    \end{scope}
  }
  \makeatother
  \tikzset{every show grid/.style={show grid/keep bb}%
  }%

\usepackage{silence}
\begin{document}
\runninghead{Chen \textit{et al.}}
\title{Adaptive Robotic Information Gathering via Non-Stationary Gaussian Processes}
\author{Weizhe Chen\affilnum{1}, Roni Khardon\affilnum{1}, and Lantao Liu\affilnum{1}}
\affiliation{\affilnum{1}Luddy School of Informatics, Computing, and Engineering, Indiana University, Bloomington, IN, USA, 47405.}
\corrauth{Lantao Liu}
\email{lantao@iu.edu}
\begin{abstract}
Robotic Information Gathering~(RIG) is a foundational research topic that answers how a robot~(team) collects informative data to efficiently build an accurate model of an unknown target function under robot embodiment constraints.
RIG has many applications, including but not limited to autonomous exploration and mapping, 3D reconstruction or inspection, search and rescue, and environmental monitoring.
A RIG system relies on a probabilistic model's prediction uncertainty to identify critical areas for informative data collection.
Gaussian Processes~(GPs) with \textit{stationary} kernels have been widely adopted for spatial modeling.
However, real-world spatial data is typically \textit{non-stationary} -- different locations do not have the same degree of variability.
As a result, the prediction uncertainty does not accurately reveal prediction error, limiting the success of RIG algorithms.
We propose a family of non-stationary kernels named Attentive Kernel~(AK), which is simple, robust, and can extend any existing kernel to a non-stationary one.
We evaluate the new kernel in elevation mapping tasks, where AK provides better accuracy and uncertainty quantification over the commonly used stationary kernels and the leading non-stationary kernels.
The improved uncertainty quantification guides the downstream informative planner to collect more valuable data around the high-error area, further increasing prediction accuracy.
A field experiment demonstrates that the proposed method can guide an Autonomous Surface Vehicle~(ASV) to prioritize data collection in locations with significant spatial variations, enabling the model to characterize salient environmental features.
\end{abstract}
\maketitle
\section{Introduction}\label{sec:1}
Collecting informative data for effective modeling of an unknown physical process or phenomenon has been studied in different domains, \textit{e.g.}, Optimal Experimental Design in Statistics~\citep{atkinson1996usefulness}, Optimal Sensor Placement in Wireless Sensor Networks~\citep{krause2008near},  Active Learning~\citep{settles2012active} and Bayesian Optimization~\citep{snoek2012practical} in Machine Learning.

In Robotics, this problem falls within the spectrum of \emph{Robotic Information Gathering~(RIG)}~\citep{thrun2002probabilistic}.
RIG has recently received increasing attention due to its wide applicability.
Applications include environmental modeling and monitoring~\citep{dunbabin2012robots}, 3D reconstruction and inspection~\citep{hollinger2013active,schmid2020efficient}, search and rescue~\citep{meera2019obstacle}, exploration and mapping~\citep{ghaffari2019sampling}, as well as active System Identification~\citep{buisson2020actively}.

\begin{figure}[tbp]
    \centering
    \includegraphics[width=0.9\linewidth]{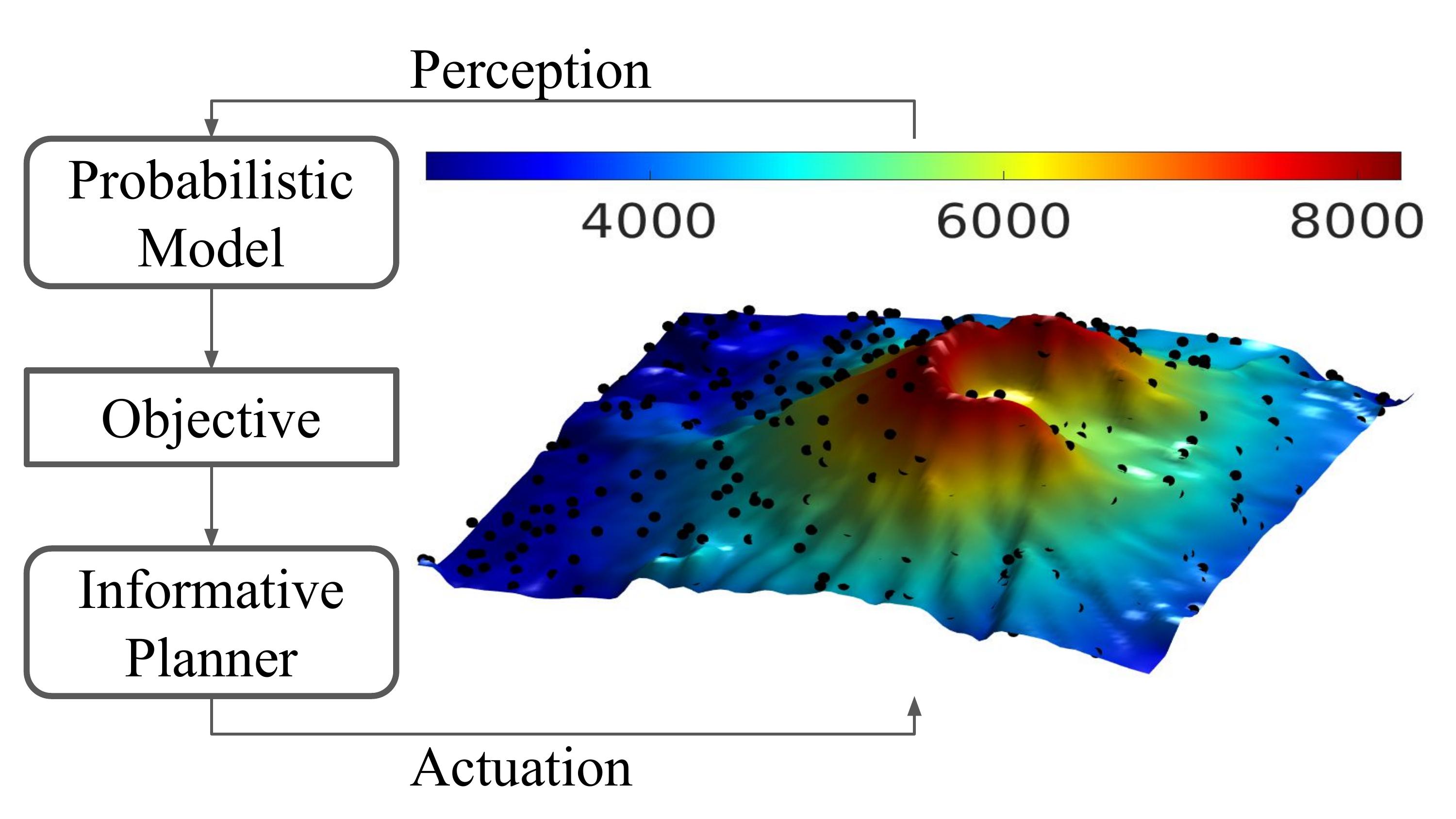}
    \caption{\textbf{Diagram of A Robotic Information Gathering System}. The goal is to autonomously gather informative elevation measurements of Mount St. Helens to efficiently build a terrain map unknown \textit{a priori}. The color indicates elevation, and black dots are collected samples.}\label{fig:volcano_env}%
\end{figure}

A RIG system typically relies on a probabilistic model’s prediction uncertainty to identify critical areas for informative data collection.
\Cref{fig:volcano_env} illustrates the workflow of a RIG system, which shows  three major forces that drive the progress of RIG: probabilistic models, objective functions, and informative planners.

\input{figure_2_motivation.tex}

The defining element distinguishing other \textit{active information acquisition} problems and RIG is the robot embodiment's physical constraints~\citep{taylor2021active}.
In Active Learning~\citep{biyik2020active} or Optimal Sensor Placement~\citep{krause2008near}, an agent can sample arbitrary data in a given space.
In RIG, however, a robot must collect data sequentially along the motion trajectories.
Consequently, most existing work in RIG is dedicated to a sequential decision-making problem called \emph{Informative (Path) Planning}~\citep{binney2013optimizing,hollinger2014sampling,lim2016adaptive,choudhury2018data,ghaffari2019sampling,best2019dec}.
Specifically, Informative Planning seeks an action sequence or a policy by optimizing an objective function that guides the robot to collect informative data, aiming to efficiently build an accurate model of the process under the robot's motion and sensing cost constraints~\citep{chen2019pareto,popovic2020informative}.
The decisive objective function is derived from the uncertainty of probabilistic models such as Gaussian processes (GPs)~\citep{ghaffari2018gaussian}, Hilbert maps~\citep{senanayake2017bayesian}, occupancy grid maps~\citep{charrow2015information}, and Gaussian mixture models~\citep{dhawale2020efficient}.
Since the performance of a RIG system depends on not only planning but also learning, as shown in the feedback loop of~\Cref{fig:volcano_env}, a natural question is: how can we further boost the performance by improving the probabilistic models?
In this work, we answer this question from the perspective of improving the \emph{modeling flexibility} and \emph{uncertainty quantification} of GPs.

Gaussian Process Regression (GPR) is one of the most prevalent methods for mapping continuous spatiotemporal phenomena.
GPR requires the specification of a kernel, and \emph{stationary} kernels, \textit{e.g.}, the radial basis function~(RBF) kernel and the Mat\'ern family, are commonly adopted~\citep{rasmussen2005mit}.
However, real-world spatial data typically does not satisfy stationary models which assume different locations have the same degree of variability.
For instance, the environment in \Cref{fig:volcano_env} shows higher spatial variability around the crater.
Due to the mismatch between the assumption and the ground-truth environment, GPR with stationary kernels cannot portray the characteristic environmental features in detail.
\Cref{fig:volcano_rbf_prediction} shows the over-smoothed prediction of the elevation map after training a stationary GPR using the collected data shown in \Cref{fig:volcano_env}.
The model also assigns low uncertainty to the high-error area, \textit{c.f.}, the circled regions in \Cref{fig:volcano_rbf_uncertainty} and \Cref{fig:volcano_rbf_error}, leading to degraded performance when the model is used in RIG.

Non-stationary GPs, on the other hand, are of interest in many applications, and the past few decades have witnessed great advancement in this research field~\citep{gibbs1997bayesian,pacriorek2003nonstationary,lang2007adaptive,plagemann2008nonstationary,plagemann2008learning,wilson2016deep,calandra2016manifold,heinonen2016nonstationary,remes2017nonstationary,remes2018neural}.
However, prior work leaves room for improvement.
The problem is that many non-stationary models learn fine-grained variability at every location, making the model too flexible to be trained without advanced parameter initialization and regularization techniques.
We propose a family of non-stationary kernels named \emph{Attentive Kernel}~(AK) to mitigate this issue.
The main idea of our AK is limiting the non-stationary model to combine a fixed set of correlation scales, \textit{i.e.}, primitive length-scales, and mask out data across discontinuous jumps by ``soft'' selection of relevant data.
The correlation-scale composition and data selection mechanisms are learned from data.
\Cref{fig:volcano_ak_prediction} shows the prediction of GPR with the AK on the same dataset used in \Cref{fig:volcano_rbf_prediction}.
As the arrows highlight, the AK depicts the environment at a finer granularity.
\Cref{fig:volcano_ak_uncertainty} and \Cref{fig:volcano_ak_error} show that the AK allocates high uncertainty to the high-error area; thus, sampling the high-uncertainty locations can help the robot collect valuable data to decrease the prediction error further.

\subsection{Contributions}
The main contribution of this paper is in designing the Attentive Kernel~(AK) and evaluating its suitability for Robotic Information Gathering~(RIG).
We present an extensive evaluation to compare the AK with existing non-stationary kernels and a stationary baseline.
The benchmarking task is elevation mapping in several natural environments that exhibit a range of non-stationary features.
The results reveal a significant advantage of the AK when it is used in passive learning,  active learning, and RIG.
We also conduct a field experiment to demonstrate the behavior of the proposed method in a real-world elevation mapping task, where the prediction uncertainty of the AK guides an Autonomous Surface Vehicle~(ASV) to identify essential sampling locations and collect valuable data rapidly.
Last but not least, we release the code (\href{https://github.com/weizhe-chen/attentive_kernels}{github.com/weizhe-chen/attentive\_kernels}) for reproducing all the results.

This paper presents an extended and revised version of previous work by \cite{chen2022ak}.
The major modifications include a comprehensive literature review on RIG to contextualize our work, additional evaluation, results, and discussion on the AK, and a substantially improved Python library.
Specifically, we provide the following contributions:
\begin{itemize}
    \item We present a broader and deeper survey on related work to highlight how our work fits into the existing literature on RIG.
    \item We add more results to the experiments and discuss them in detail to provide further evidence for our conclusions.
    \item We thoroughly evaluate the AK from various perspectives and discuss its limitations and potential future work.
    \item We release a new Python library called PyPolo (\href{https://pypolo.readthedocs.io/}{pypolo.readthedocs.io}) for learning, researching, and benchmarking RIG algorithms. This library is a significant improvement and restructure compared to the one presented in \cite{chen2022ak}.
\end{itemize}

\section{Related Work}\label{sec:2}
In this section, we will first survey related work in RIG, which mainly revolves around three pillars: probabilistic models (\Cref{sec:rig_models}), objective functions (\Cref{sec:rig_objectives}), and informative planning algorithms (\Cref{sec:rig_planning}).
Also, we discuss relevant RIG applications in~\Cref{sec:related_applications}.
Then, we categorize prior efforts on non-stationary GPs and how the proposed method relates to the existing solutions (\Cref{sec:nonstat}).
Finally, we describe the relationship between RIG and some related research topics to locate our work within the context of existing literature (\Cref{sec:related_topics})

\subsection{Robotic Information Gathering}\label{sec:related_rig}
A RIG system has three essential components:
\begin{enumerate}
    \item A model to approximate the unknown target function;
    \item An objective function that can characterize the model's prediction error;
    \item An informative planner that makes \emph{non-myopic} decisions by optimizing the objective function under the robot's embodiment constraints.
\end{enumerate}
We discuss these three aspects in this section.

\subsubsection{Objective Functions}\label{sec:rig_objectives}
RIG can be the main goal of some tasks, such as infrastructure inspection~\citep{bircher2018receding}, or serve as an auxiliary task for achieving other goals, \textit{e.g.}, seeking the biological hotspots in an unknown environment~\citep{mccammon2018topological}.
In the former cases, the objective function is purely ``\emph{information-driven}''~\citep{ferrari2021information,bai2021information}, while in the latter scenarios, the objective function balances exploration and exploitation~\citep{marchant2012bayesian,marchant2014bayesian,bai2016information}.
The objective function can be further extended to multi-objective cases~\citep{chen2019pareto,ren2022local,dang2020resilient}.

Many objective functions have been proposed, inspired by Information Theory and Optimal Experimental Design~\citep{charrow2015information,zhang2020fsmi,carrillo2015monotonicity}.
Information-theoretic objective functions include Shannon's and R\'enyi's entropy, mutual information, and Kullback-Leibler divergence between the prior and posterior predictive distributions.
In the case of multivariate Gaussian distributions, these information measures are all related to the logarithmic determinant of the posterior covariance matrix, which can be intuitively viewed as computing the ``size'' of the posterior covariance matrix.
Optimal design theory directly measures the size by computing the matrix determinant, trace, or eigenvalues.
Computing the matrix determinant and eigenvalue is known to be computationally expensive.
Therefore, many existing works on objective functions are dedicated to alleviating the computational bottleneck~\citep{charrow2015icra,charrow2015information,zhang2020fsmi,zhang2020fisher,gupta2021efficient,xu2021crmi}.

Most objective functions are summary statistics of the predictive (co)variance given by a probabilistic model.
Only when the predictive (co)variance captures modeling error well, optimizing these objective functions can guide the robot to collect informative data that effectively improve the model's accuracy.
From this perspective, improving the uncertainty-quantification capability of probabilistic models can broadly benefit future work based on these objective functions.
This aspect is what we strive to improve in this work.
As  can be seen in the next section, this problem is understudied.

\subsubsection{Probabilistic Models}\label{sec:rig_models}
Many probabilistic models have been applied to RIG, \textit{e.g.}, Gaussian processes~\citep{stachniss2009learning,marchant2012bayesian,marchant2014bayesian,ouyang2014multi,ma2017informative,luo2018adaptive,jang2020multi,popovic2020localization,lee2022trust}, Hilbert maps~\citep{ramos2016hilbert,senanayake2017bayesian,guizilini2019variational}, occupancy grid maps~\citep{popovic2017online,popovic2020informative,saroya2021roadmap}, and Gaussian mixture models~\citep{meadra2018variable,tabib2019real}.
GPs are widely adopted due to their excellent uncertainty quantification feature, which is decisive to RIG.
However, the vanilla GP models need to be more computationally efficient to be suitable for real-time applications and multi-robot scenarios.
Therefore, related work in RIG mainly discusses GPs in the context of improving computational efficiency and coordinating multiple robots.
\cite{jang2020multi} apply the distributed GPs~\citep{deisenroth2015distributed} to decentralized multi-robot online Active Sensing.
\cite{ma2017informative} and \cite{stachniss2009learning} use sparse GPs to alleviate the computational burden.
The mixture of GP experts~\citep{rasmussen2001infinite} has been applied to divide the workspace into smaller parts for multiple robots to model an environment simultaneously~\citep{luo2018adaptive,ouyang2014multi}.

The early work by \cite{krause2007nonmyopic} is highly related to our work.
They use a spatially varying linear combination of localized stationary processes to model the non-stationary pH values in a river.
The weight of each local GP is the normalized predictive variance at the test location.
This idea is similar to the length-scale selection idea in \Cref{sec:ls_selection}.
The main difference is that they manually partition the workspace while our model learns a weighting function from data.
To the best of our knowledge, our work is the first to discuss the influence of the probabilistic models' uncertainty quantification on RIG performance.

\subsubsection{Informative Planning}\label{sec:rig_planning}
The problem of seeking an action sequence or policy that yields informative data is known as Informative \emph{Path} Planning due to historical reasons~\citep{singh2007efficient,meliou2007nonmyopic}.
However, the problem is not restricted to path planning.
For example, recent work has discussed informative \emph{motion} planning~\citep{teng2021toward}, informative \emph{view} planning~\citep{lauri2020multi}, and exploratory \emph{grasping}~\citep{danielczuk2021exploratory}.
Hence, we adopt the generic term Informative Planning to unify different branches of the same problem.

Early works on Informative Planning propose various \textit{recursive greedy} algorithms that provide performance guarantee by exploiting the \textit{submodularity} property of the objective function~\citep{singh2007efficient,meliou2007nonmyopic,binney2013optimizing}.
Note that the performance guarantee is on uncertainty reduction rather than modeling accuracy.
Planners based on dynamic programming~\citep{low2009information,cao2013multi} and mixed integer quadratic programming~\citep{yu2014correlated} lift the assumption on the objective function at the expense of higher computational complexity.
These methods solve combinatorial optimization problems in discrete domains, thus scaling poorly in problem size.
To develop efficient planners in \textit{continuous} space with motion constraints, \cite{hollinger2014sampling} introduce sampling-based informative motion planning, which is further developed to online variants~\citep{schmid2020efficient, ghaffari2019sampling}.
Monte Carlo Tree Search~(MCTS) methods are conceptually similar to sampling-based informative planners~\citep{kantaros2021sampling,schlotfeldt2018anytime} and have recently garnered great attention~\citep{arora2019multimodal,best2019dec,morere2017sequential,chen2019pareto,flaspohler2019information}.
Trajectory optimization is a solid competitor to sampling-based planners.
Bayesian Optimization~\citep{marchant2012bayesian,bai2016information,di2021multi} and Evolutionary Strategy~\citep{popovic2017online,popovic2020informative,hitz2017adaptive} are the two dominating methods in this realm.
New frameworks of RIG, \textit{e.g.}, Imitation Learning~\citep{choudhury2018data}, are emerging.
Communication constraints~\citep{lauri2017multi} and adversarial attacks~\citep{schlotfeldt2021resilient} have also been discussed.

\subsubsection{Relevant Applications}\label{sec:related_applications}
Mobile robots can be considered as autonomous data-gathering tools, enabling scientific research in remote and hazardous environments~\citep{li2020exploration,bai2021information}.
RIG has been successfully applied to environmental mapping and monitoring~\citep{dunbabin2012robots}.
An underwater robot with a profiling sonar can inspect a ship hull autonomously~\citep{hollinger2013active}.
In \cite{girdhar2014autonomous}, the underwater robot performs semantic exploration with online topic modeling, which can group corals belonging to the same species or rocks of similar types.
\cite{flaspohler2019information} deploy an ASV for localizing and collecting samples at the most exposed coral head.
\cite{hitz2017adaptive} monitor algal bloom using an ASV, which can provide early warning to environmental managers to conduct water treatment in a more appropriate time frame.
\cite{manjanna2018heterogeneous} show that a robot team can help scientists collect plankton-rich water samples via \textit{in situ} mapping of Chlorophyll density.
\cite{fernandez2022informative} propose delineating the sampling locations that correspond to the quantile values of the phenomenon of interest, which helps the scientists to collect valuable data for later analysis.
Active lakebed mapping, where the static ground truth is available, can serve as a testbed for ocean bathymetric mapping~\citep{ma2018data}.
RIG can also be applied to the 3D reconstruction of large scenes~\citep{kompis2021informed} and object surfaces~\citep{zhu2021online}.
In addition to geometric mapping, semantic mapping is also explored in \citep{atanasov2014nonmyopic}, where a PR2 robot with an RGB-D camera attached to the wrist leverages non-myopic view planning for active object classification and pose estimation.
\cite{meera2019obstacle} present a realistic simulation of a search-and-rescue scenario in which informative planning maximizes search efficiency under the Unmanned Aerial Vehicle~(UAV) flight time constraints.
Fixed-wing UAVs use aerodynamics akin to aircraft, so it has a much longer flight time than multi-rotors.
\cite{moon2022tigris} simulate a fixed-wing UAV with a forward-facing camera to search for multiple objects of interest in a large search space.

\subsection{Non-Stationary Gaussian Processes}\label{sec:nonstat}
GPs suffer from two significant limitations~\citep{rasmussen2001infinite}.
The first one is the notorious cubic computational complexity of a vanilla implementation.
Recent years have witnessed remarkable progress in solving this problem based on sparse GPs~\citep{quinonero2005unifying,titsias2009variational,hoang2015unifying,sheth2015sparse,bui2017unifying,wei2021direct}.
The second drawback is that the covariance function is commonly assumed to be stationary, limiting the modeling flexibility.
Developing non-stationary GP models that are easy to train is still an active open research problem.
Ideas of handling non-stationarity can be roughly grouped into three categories: input-dependent length-scale~\citep{gibbs1997bayesian,pacriorek2003nonstationary,lang2007adaptive,plagemann2008learning,plagemann2008nonstationary,heinonen2016nonstationary,remes2017nonstationary}, input warping~\citep{sampson92nonparametric,snoek2014input,calandra2016manifold,wilson2016deep,tompkins2020sparse,salimbeni2017doubly}, and the mixture of experts~\citep{rasmussen2001infinite,trapp2020deep}.

Input-dependent length-scale provides excellent flexibility to learn different correlation scales at different input locations.
\cite{gibbs1997bayesian} and \cite{pacriorek2003nonstationary} have shown how one can construct a valid kernel with input-dependent length-scales, namely, a \emph{length-scale function}.
The standard approach uses another GP to model the length-scale function, which is then used in the kernel of a GP, yielding a hierarchical Bayesian model.
Several papers have developed inference techniques for such models and demonstrated their use in some applications~\citep{lang2007adaptive,plagemann2008learning,plagemann2008nonstationary,heinonen2016nonstationary,remes2017nonstationary}.
Recently, \cite{remes2018neural} show that modeling the length-scale function using a neural network improves performance.
Note, however, that learning a length-scale function is nontrivial~\citep{wang2020nonseparable}.

Input warping is more widely applicable because it endows any stationary kernel with the ability to model non-stationarity by mapping the input locations to a distorted space and assuming stationarity holds in the new space.
This approach has a tricky requirement: the mapping must be \emph{injective} to avoid undesirable folding of the space~\citep{sampson92nonparametric,snoek2014input,salimbeni2017doubly}.

A mixture of GP experts~(MoGPE) uses a \emph{gating network} to allocate each data to a local GP that learns its hyper-parameters from the assigned data.
It typically requires Gibbs sampling~\citep{rasmussen2001infinite}, which can be slow.
Hence, one might need to develop a faster approximation~\citep{nguyen2008local}.
We view MoGPE as an orthogonal direction to other non-stationary GPs or kernels because any GP model can be treated as the expert so that one can have a mixture of non-stationary GPs.

The AK lies at the intersection of these three categories.
\Cref{sec:ls_selection} presents an input-dependent length-scale idea by weighting base kernels with different fixed length-scales at each location.
Composing base kernels reduces the difficulty of learning a length-scale function from scratch and makes our method compatible with any base kernel.
In \Cref{sec:instance_selection}, we augment the input with extra dimensions.
We can view the augmentation as warping the input space to a higher-dimensional space, ensuring \emph{injectivity} by design.
Combining these two ideas gives a conceptually similar model to MoGPE~\citep{rasmussen2001infinite} in that they both divide the space into multiple regions and learn localized hyper-parameters.
The idea of augmenting the input dimensions has been discussed by \cite{pfingsten2006nonstationary}.
However, they treat the augmented vector as a latent variable and resort to Markov chain Monte Carlo for inference.
The AK treats the augmentation vector as the output of a deterministic function of the input, resulting in a more straightforward inference procedure.
Also, the AK can be used in MoGPE to build more flexible models.

In robotic mapping, another line of notable work on probabilistic models is the family of Hilbert maps~\citep{ramos2016hilbert,senanayake2017bayesian,guizilini2019variational}, which aims to alleviate the computational bottleneck of GPs~\citep{callaghan2012Gaussian} by projecting the data to another feature space and applying a logistic regression classifier in the new space.
Since Hilbert maps are typically used for occupancy mapping~\citep{doherty2016probabilistic} and reconstruction tasks~\citep{guizilini2017learning}, related work also considers non-stationarity for better prediction~\citep{senanayake2018automorphing,tompkins2020online}.

\begin{figure}[tbp]
    \centering
    \includegraphics[width=\linewidth]{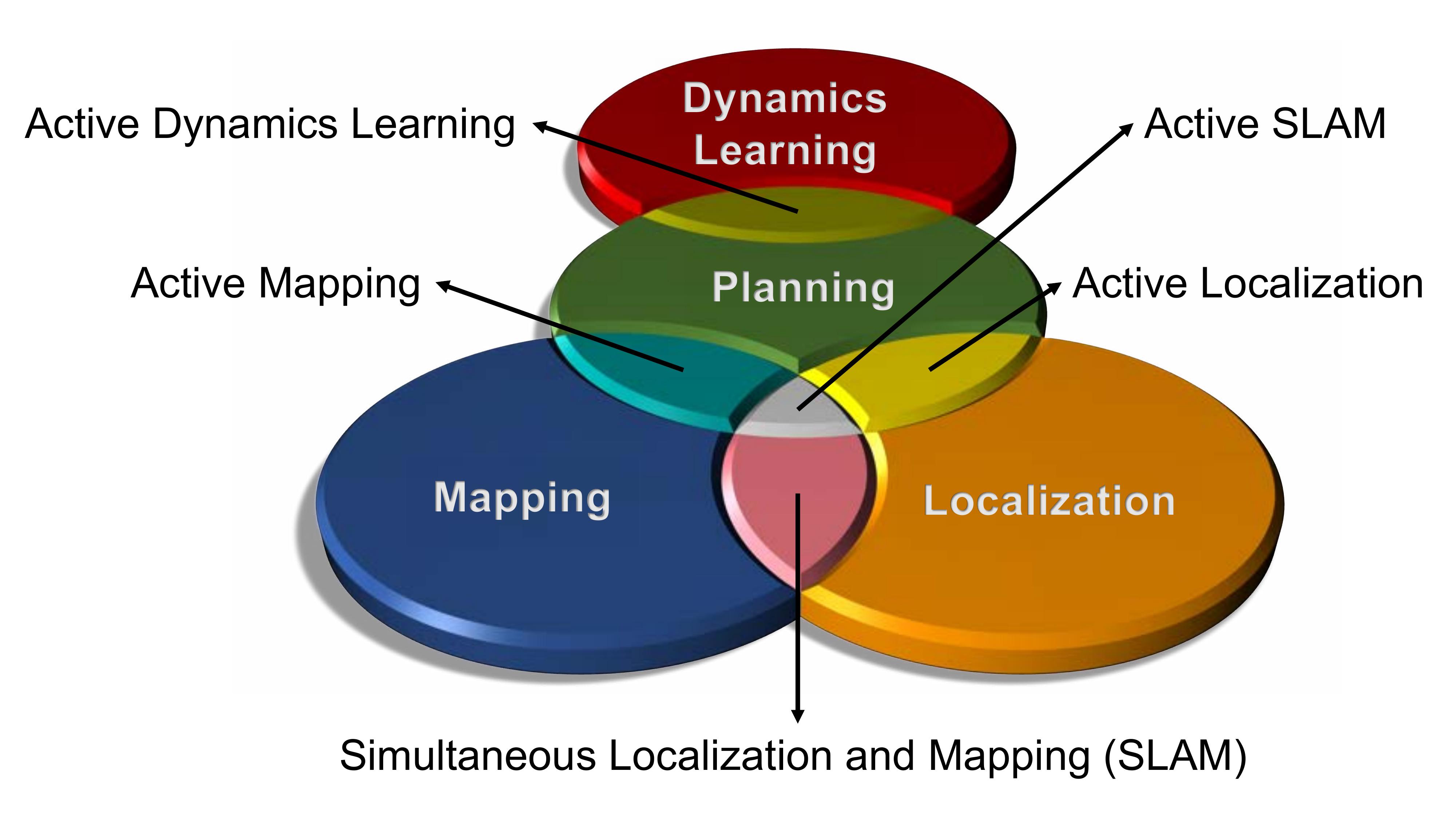}
    \caption{\textbf{Research Topics Related to RIG}.}\label{fig:venn_diagram}%
\end{figure}

\subsection{Relationship to Other Research Topics}\label{sec:related_topics}
RIG is a fundamental research problem seeking an answer to the following question:

\textit{How does a robot (team) collect informative data to efficiently build an accurate model of an unknown function under robot embodiment constraints?}

Depending on how we define \emph{data}  and what the unknown \emph{target function} is, RIG appears in the form of Active Dynamics Learning, Active Mapping, Active Localization, and Active Simultaneous Localization and Mapping~(SLAM).
\Cref{fig:venn_diagram} shows a Venn diagram of these topics.
Although we evaluate the AK in Active Mapping tasks, other related problems, \textit{e.g.}, Active Dynamics Learning, can also benefit from the proposed method if the target function is modeled by a GP.
On top of that, guiding the data collection process by minimizing \emph{well-calibrated} uncertainty estimates applies to all these related topics~\citep{rodriguez2018importance}.

\subsubsection{Active Dynamics Learning}\label{sec:active_sysid}
Control synthesis typically depends on the system dynamics.
Due to the complex interaction between the robot and the environment, \textit{e.g.}, a quadruped running at high speed over rough terrain, mechanical wear and tear, and actuator faults, it may be infeasible to build an accurate dynamics model \textit{a priori}~\citep{cully2015robots}.
In these cases, the robot must take safe actions and observe its dynamics to explore different behavioral regimes sample-efficiently~\citep{abraham2019active}.
When the robot collects dynamics information to infer the unknown transition function, the RIG problem is known as Active Dynamics Learning or System Identification~\citep{taylor2021active}.
In this context, informative \emph{data} refers to the state-action-state pairs or the full state-action trajectories that help efficiently learn an accurate model of the unknown \emph{system dynamics} or \emph{transition function}.
The system dynamics can be modeled as fixed-form equations~\citep{jegorova2020adversarial}, data-driven models, including \emph{parametric} models~\citep{chua2018deep}, \emph{non-parametric} models~\citep{calandra2016manifold}, and \emph{semi-parametric} models~\citep{romeres2019semiparametrical}, and the combination of the analytical models and data-driven models~\citep{heiden2021neuralsim}.
GPs have arguably become the \textit{de facto} standard in collecting informative data that minimizes the predictive uncertainty of data-driven models to achieve sample-efficient dynamics learning~\citep{rezaei2019cascaded,buisson2020actively,capone2020localized,lew2022safe,yu2021active}.
With the rise of Automatic Differentiation~\citep{pytorch}, a large body of recent work tend to estimate the physical parameters inside \emph{differentiable Rigid-Body Dynamics} models~\citep{sutanto2020encoding,lutter2021differentiable,de2018end} or \emph{differentiable robotics simulators}~\citep{hu2019chainqueen,freeman2021brax, werling2021fast}.
The literature emphasizes that calibrating the simulation~\citep{mehta2021user} is essential for both Reinforcement Learning with domain randomization~\citep{ramos2019bayessim,muratore2022robot} or trajectory optimization~\citep{du2021underwater,heiden2021disect}.
In this context, we can consider RIG as Active Simulation Calibration since the robot collects informative trajectories to efficiently learn an accurate model of the unknown simulation parameters under the kinodynamic constraints.
Active Simulation Calibration can also directly optimize the task-specific reward.
For instance, \cite{muratore2021data} model the policy return as a GP and use Bayesian Optimization to tune the simulation parameters.
\cite{liang2020learning} learn a task-oriented exploration policy to collect informative data for calibrating task-relevant simulation parameters.

\subsubsection{Active Perception}\label{sec:active_perception}
When the robot collects data from the \emph{environment} rather than its dynamics, RIG becomes \emph{Active Perception} -- an agent (\textit{e.g.}, camera or robot) changes its angle of view or position to perceive the surrounding environment better~\citep{bajcsy1988active,aloimonos1988active,bajcsy2018revisiting}.
If the agent actively perceives the environment to reduce the \emph{localization uncertainty}, the problem is referred to as Active Localization~\citep{fox1998active,borghi1998minimum}.
If the goal is to build the best possible \emph{representation of an environment}, the problem essentially becomes Active Mapping~\citep{lluvia2021active}.

\subsubsection{Active Localization}\label{sec:active_localization}
Localization uncertainty can arise from perceptual degradation~\citep{ebadi2020lamp}, noisy actuation~\citep{thrun2002probabilistic}, and inaccurate modeling~\citep{roy1999coastal}.
Decision-making or planning under uncertainty~\citep{lavalle2006planning,bry2011rapidly,preston2022robotic} provides an elegant framework to formulate these problems using partially observable Markov decision processes~(POMDP)~\citep{kaelbling1998planning,cai2021hyp,lauri2022partially}.
A principled approach to address these problems is to plan in the \emph{belief space}~\citep{kaelbling2013integrated,nishimura2021sacbp}.
\emph{Information gathering} is a natural behavior generated by Belief-Space Planning~\citep{platt2010belief}.
Computing optimal policy in belief space is computationally intensive, but useful heuristics enable efficient computation of high-quality solutions~\citep{kim2019pomhdp,prentice2009belief,zheng2022belief}.
Although the localization uncertainty can come from different sources, in perceptually degraded environments such as subterranean, perception uncertainty outweighs the others.
A dedicated topic for this case is perception-aware planning~\citep{zhang2020active}.
Note that localization is not necessarily positioning a mobile robot on a map~\citep{chaplot2018active}; it can also be locating and tracking an object in the workspace of a manipulator with force-torque sensor measurements~\citep{wirnshofer2020controlling,schneider2022active}.

\subsubsection{Active Sensing and Mapping}\label{sec:active_mapping}
Suppose the data refers to the robot's observations, \textit{e.g.}, camera images or LiDAR point clouds, and the unknown target function is the ground-truth representation of the environment.
In that case, RIG can be considered an Active Mapping problem~\citep{placed2022survey}.
Mapping uncertainty can come from \textit{aleatoric} uncertainty inherent in measurement noise and \textit{epistemic} uncertainty due to unknown model parameters and data scarcity~\citep{krause2007nonmyopic}.
Active Mapping efficiently builds an accurate model of the environment by minimizing epistemic uncertainty, which is often termed Active Sensing when focusing on the active acquisition of sensor measurements for better \emph{prediction} rather than \emph{model learning}~\citep{cao2013multi,macdonald2019active,schlotfeldt2019maximum,ruckin2022adaptive}.
When mapping a 3D environment using a sensor with a limited field-of-view, this is known as the Next-Best View problem~\citep{connolly1985determination,bircher2016receding,palomeras2019autonomous,lauri2020multi}.
Autonomous Exploration is sometimes used interchangeably with Active Mapping~\citep{lluvia2021active}.
However, the nuances of the assumptions and evaluation metrics of the two domains yield significantly different solutions and robot behaviors.
Specifically, Active Mapping typically assumes ideal localization~\citep{popovic2020localization} and aims at building an accurate environment map using noisy and sparse observations; thus, the performance is evaluated by reconstruction error against the ground truth.
The robot might revisit some complex regions to collect more data if the model prediction is not accurate enough.
For example, when performing Active Mapping of a ship hull, the robot should collect more data around the propeller~\citep{hollinger2013active}.
Autonomous Exploration emphasizes obtaining the global structure of a vast unknown environment, implying that the robot (team) should avoid duplicate coverage; thus, the evaluation criterion is the explored volume~\citep{cao2021tare}.
In contrast to Active Mapping, unreliable localization is one of the major challenges in Autonomous Exploration that should be addressed~\citep{tranzatto2022team,papachristos2019localization}.
In this work, our application belongs to the Active Mapping problem, where the better uncertainty quantification of the proposed non-stationary GPR guides the robot to collect more informative data for rapid learning of an accurate map.

\subsubsection{Active SLAM}\label{sec:active_slam}
Controlling a robot performing SLAM to reduce both the localization and mapping uncertainty is called active SLAM~\citep{placed2022survey}.
Active Localization and Active Mapping are two conflicting objectives.
The former asks the robot to revisit explored areas for potential \emph{loop closure}~\citep{stachniss2004exploration}, while the latter guides the robot to expand \emph{frontiers} for efficient map building~\citep{yamauchi1997frontier}.
We refer the interested reader to the corresponding survey papers~\citep{lluvia2021active,placed2022survey}.

\rowcolors{2}{gray!25}{white}
\begin{table}[tbp]
  \caption{Mathematical Notations.}\label{tab:notations}
  \centering
  \small
  \begin{tabular}{lll}
    \toprule
    {Meaning}           & {Example}           & {Remark}            \\
    \midrule
    variable            & $m$                 & lower-case          \\
    constant            & $M$                 & upper-case          \\
    vector              & $\mathbf{x}$        & bold, lower-case    \\
    matrix              & $\mathbf{X}$        & bold, upper-case    \\
    set/space           & $\mathbb{R}$        & blackboard          \\
    Cartesian product   & $[a,b]^{D}$         & $D$-dim hypercube   \\
    function            & $\mathtt{d}(\cdot)$ & typewriter          \\
    special PDF         & $\mathcal{N}$       & calligraphy capital \\
    definition          & $\triangleq$        & normal              \\
    transpose           & $\mathbf{m}^{\T}$   & customized command  \\
    Euclidean norm      & $\norm{\cdot}$      & customized command  \\
    \bottomrule
  \end{tabular}%
\end{table}

\section{Problem Statement}\label{sec:problem_statement}
Consider deploying a robot to \emph{efficiently} build a map of an \emph{unknown} environment using only \emph{sparse} sensing measurements of onboard sensors.
For instance, when reconstructing a pollution distribution map, the environmental sensors can only measure the pollutant concentration in a \emph{point-wise} sampling manner, yielding sparse measurements along the trajectory.
Another scenario is building a large bathymetric map of the seabed.
The depth measurements of a multi-beam sonar can be viewed as \emph{point measurements} because the unknown target area is typically vast.
Exhaustively sampling the whole environment is prohibitive, if not impossible; thus, one must develop adaptive planning algorithms to collect the most informative data for building an accurate model.
\Cref{tab:notations} introduces the notation system used in this paper.
We use column vector by default.

\subsection{Minimization of Error \textit{vs.} Uncertainty}
\begin{problem}\label{prob:1}%
    The target environment is an unknown function $\mathtt{f}_{\text{env}}(\mathbf{x}):\real^{D}\mapsto\real$ defined over spatial locations $\mathbf{x}\in\real^{D}$.
    Let $\mathbb{T}\triangleq\{t\}_{t=0}^{T}$ be the set of decision epochs.
    A robot at state $\mathbf{s}_{t-1}\in\mathbb{S}$ takes an action $a_{t-1}\in\mathbb{A}$, arrives at the next state $\mathbf{s}_{t}$ following a transition function $p(\mathbf{s}_{t}\mid\mathbf{s}_{t-1},a_{t-1})$, and collects $N_{t}\in\mathbb{N}$ noisy measurements $\mathbf{y}_{t}\in\real^{N_{t}}$ at sampling locations $\mathbf{X}_{t}=[\mathbf{x}_{1},\dots,\mathbf{x}_{N_{t}}]^{\T}\in\real^{N_{t}\times{D}}$ when transitioning from $\mathbf{s}_{t-1}$ to $\mathbf{s}_{t}$.
    We assume that the transition function is known and deterministic and that the robot state is observable.
    The robot maintains a probabilistic model built from all the training data collected so far $\mathbb{D}_{t}=\{(\mathbf{X}_{i},\mathbf{y}_{i})\}_{i=1}^{t}$.
    The model provides predictive mean $\mathtt{\mu}_{t}:\real^{D}\mapsto\real$ and predictive variance $\mathtt{\nu}_{t}:\real^{D}\mapsto\real_{\geq{0}}$ functions.
    Let $\mathbf{x}^{\star}$ be a test or query location, and $\mathtt{error}(\cdot)$ be an error metric.
    At each decision epoch $t\in\mathbb{T}$, our goal is to find sampling locations that minimize the \emph{expected error} after updating the model with the collected data%
    \begin{equation}\label{eq:problem_error}%
        \argmin_{\mathbf{X}_{t}}\mathtt{E}_{\mathbf{x}^{\star}}\left[\mathtt{error}\left(\mathtt{f}_{\text{env}}(\mathbf{x}^{\star}),\mathtt{\mu}_{t}(\mathbf{x}^{\star}),\mathtt{\nu}_{t}(\mathbf{x}^{\star})\right)\right].%
    \end{equation}%
\end{problem}%
The predictive variance is also included in \Cref{eq:problem_error} because it is required when computing some error metrics, \textit{e.g.}, negative log predictive density.
Note that the expected error cannot be directly used as the objective function for a planner because the ground-truth function $\mathtt{f}_{\text{env}}$ is unknown.
RIG bypasses this problem by optimizing a surrogate objective.

\begin{problem}\label{prob:2}%
    Assuming the same conditions as \Cref{prob:1}, find \emph{informative} sampling locations that minimize an uncertainty measure $\mathtt{info}(\cdot)$, \textit{e.g.,} entropy:
    \begin{equation}\label{eq:problem_info}%
        \argmin_{\mathbf{X}_{t}}\mathtt{E}_{\mathbf{x}^{\star}}\left[\mathtt{info}\left(\mathtt{\nu}_{t}(\mathbf{x}^{\star})\right)\right].%
    \end{equation}
\end{problem}

RIG implicitly assumes that minimizing prediction uncertainty (\Cref{prob:2}) can also effectively reduce prediction error (\Cref{prob:1}).
This assumption is valid when the model uncertainty is \emph{well-calibrated}.
A model with well-calibrated uncertainty gives high uncertainty when the prediction error is significant and low uncertainty otherwise.

\subsection{Gaussian Process Regression}\label{sec:gpr}
The predictive mean and variance functions are given by a Gaussian process regression~(GPR) model in this work.
A Gaussian process (GP) is a collection of random variables, any finite number of which have a joint Gaussian distribution~\citep{rasmussen2005mit}.

\subsubsection{Model Specification}\label{sec:model_specification}
We place a Gaussian process \emph{prior} over the unknown target function
\begin{equation}
    \mathtt{f}_{\text{env}}(\mathbf{x})\sim\mathcal{GP}(\mathtt{m}(\mathbf{x}),\mathtt{k}(\mathbf{x},\mathbf{x}')),
\end{equation}
which is specified by a mean function $\mathtt{m}(\mathbf{x})$ and a covariance function $\mathtt{k}(\mathbf{x},\mathbf{x}')$, \textit{a.k.a.} \emph{kernel}.
After standardizing the training targets $y$ to have a near-zero mean empirically, the mean function is typically simplified to a \emph{zero function}, rendering the specification of the covariance function an important choice.
Popular choices of the covariance functions are \emph{stationary kernels} such as the RBF kernel and the Mat\'ern family.
We refer the interested reader to~\cite{rasmussen2005mit} for other commonly used kernels.

This paper uses the RBF kernel to show how we transform a stationary kernel to a non-stationary one using the proposed method.
Given two inputs $\mathbf{x}$ and $\mathbf{x}'$, the RBF kernel measures their correlation by computing the following kernel value
\begin{equation}\label{eq:rbf}
    \mathtt{k}(\mathbf{x},\mathbf{x}')=\mathtt{\exp}\left(-\frac{\norm{\mathbf{x}-\mathbf{x}'}^{2}}{2\ell^{2}}\right).
\end{equation}
The correlation scale parameter $\ell$ is called the \emph{length-scale}, which informally indicates the distance one has to move in the input space before the function value can change significantly~\citep{rasmussen2005mit}.
A given sample should be most correlated to itself; thus, \Cref{eq:rbf} gives the largest kernel value when $\mathbf{x}=\mathbf{x}'$.
Kernels are typically normalized to ensure that the largest kernel value is $1$ and an \emph{amplitude} parameter $\alpha$ can be used to scale the kernel value $\alpha\mathtt{k}(\mathbf{x},\mathbf{x}')$ to a larger range.

GPR assumes a Gaussian likelihood function.
The target values $y$ are the function outputs $f$ corrupted by an additive Gaussian white noise
\begin{equation}\label{eq:likelihood}
    \mathtt{p}(y|\mathbf{x})=\mathcal{N}(y|\mathtt{f}(\mathbf{x}),\sigma^2),
\end{equation}
where $\sigma$ is the observational \emph{noise scale}.

\subsubsection{Prediction}
Since GP is a \emph{conjugate} prior to the Gaussian likelihood, given $N$ training inputs $\mathbf{X}\in\real^{N\times{D}}$ and training targets $\mathbf{y}\in\real^{N}$, the posterior predictive distribution has a closed-form expression:
\begin{align}\label{eq:predictive_distribution}
    p(f_{\star}\rvert\mathbf{y})&=\mathcal{N}(f_{\star}\rvert\mu,\nu),\\
    \mu&=\mathbf{k}_{\star}^{\T}\mathbf{K}_{y}^{-1}\mathbf{y},\label{eq:pred_mu}\\
    \nu&=k_{\star\star}-\mathbf{k}_{\star}^{\T}\mathbf{K}_{y}^{-1}\mathbf{k}_{\star},\label{eq:pred_nu}
\end{align}
where $\mathbf{k}_{\star}$ is the vector of kernel values between all the training inputs $\mathbf{X}$ and the test input $\mathbf{x}^{\star}$, $\mathbf{K}_{y}$ is a shorthand of $\mathbf{K}_{\mathbf{x}}+\sigma^{2}\mathbf{I}$, $\mathbf{K}_{\mathbf{x}}$ is the covariance matrix given by the kernel function evaluated at each pair of training inputs, and $k_{\star\star}\triangleq\mathtt{k}(\mathbf{x}^{\star},\mathbf{x}^{\star})$.

\subsubsection{Learning}\label{sec:learning}
The prediction of GPR in~\Cref{eq:predictive_distribution} is readily available with no need to train a model.
However, the prediction quality of GPR depends on the setting of \emph{hyper-parameters} $\bm{\psi}\triangleq[\ell,\alpha,\sigma]$.
These are the parameters of the kernel and likelihood function.
Hence, optimizing these parameters -- a process known as \emph{model selection} -- is a common practice to obtain a better prediction.
Model selection is typically implemented by maximizing the \emph{model evidence}, \textit{a.k.a.}, log \emph{marginal likelihood},
\begin{equation*}\label{eq:marginal_likelihood}
    \ln{p(\mathbf{y}|\bm{\psi})}=\frac{1}{2}(\underbracket{-\mathbf{y}^{\T}\mathbf{K}_{y}^{-1}\mathbf{y}}_{\text{model fit}}-\underbracket{\ln{\mathrm{det}(\mathbf{K}_{y})}}_{\text{model complexity}}-\underbracket{N\ln(2\pi)}_{\text{constant}}),
\end{equation*}
where $\mathrm{det}(\dots)$ is the matrix determinant.

When using GPR with the commonly used stationary kernels to reconstruct a real-world environment, high uncertainty is assigned to less sampled areas, regardless of the prediction error~(see \Cref{fig:volcano_rbf_uncertainty,fig:1d_rbf}).
However, real-world spatial environments are typically non-stationary, and the high prediction error is more likely to be present in the high-variability region.
In other words, the assumption of well-calibrated uncertainty is violated when using stationary kernels.
Therefore, we aim to develop non-stationary kernels to improve GPR’s uncertainty-quantification capability and prediction accuracy.

\section{Methodology}\label{sec:4}
We propose a new kernel called Attentive Kernel to deal with non-stationarity.

\begin{definition}[Attentive Kernel (AK)]\label{def:attentive_kernel}%
  Given two inputs $\mathbf{x},\mathbf{x}'\in\real^{D}$, vector-valued functions $\mathtt{\mathbf{w}}_{\bm{\theta}}(\mathbf{x}):\real^{D}\mapsto[0,1]^{M}$ and $\mathtt{\mathbf{z}}_{\bm{\phi}}(\mathbf{x}):\real^{D}\mapsto[0,1]^{M}$ parameterized by $\bm{\theta,\phi}$, an amplitude $\alpha$, and a set of $M$ base kernels $\{\mathtt{k}_{m}(\mathbf{x},\mathbf{x}')\}_{m=1}^{M}$, let $\bar{\mathbf{w}}=\nicefrac{\mathbf{w}_{\bm{\theta}}(\mathbf{x})}{\norm{\mathtt{\mathbf{w}}_{\bm{\theta}}(\mathbf{x})}}$, and $\bar{\mathbf{z}}=\nicefrac{\mathbf{z}_{\bm{\phi}}(\mathbf{x})}{\norm{\mathtt{\mathbf{z}}_{\bm{\phi}}(\mathbf{x})}}$.
  The AK is defined as
  \begin{equation}
    \mathtt{ak}(\mathbf{x},\mathbf{x}')=\alpha\bar{\mathbf{z}}^{\T}\bar{\mathbf{z}}'\sum_{m=1}^{M}\bar{w}_{m}\bar{w}_{m}'\mathtt{k}_{m}(\mathbf{x},\mathbf{x}'),
  \end{equation}
  where $\bar{w}_{m}$ is the $m$-th element of $\bar{\mathbf{w}}$.
\end{definition}
We learn the parametric functions that map each input $\mathbf{x}$ to $\mathbf{w}$ and $\mathbf{z}$.
The weight $\bar{w}_{m}\bar{w}_{m}'$ gives \emph{similarity attention scores} to combine the set of base kernels $\{\mathtt{k}_{m}(\mathbf{x},\mathbf{x}')\}_{m=1}^{M}$.
The inner product $\bar{\mathbf{z}}^{\T}\bar{\mathbf{z}}'$ defines a \emph{visibility attention score} to mask the kernel value.

\Cref{def:attentive_kernel} is generic because any existing kernel can be the base kernel.
To address non-stationarity, we choose the base kernels to be a set of stationary kernels with the same functional form but different length-scales.
Specifically, we use RBF kernels with $M$ length-scales $\{\ell_{m}\}_{m=1}^{M}$ that are evenly spaced in the interval $[\ell_{\text{min}},\ell_{\text{max}}]$:
\begin{equation*}
  \mathtt{k}_{m}(\mathbf{x},\mathbf{x}')\triangleq\mathtt{k}_{\text{RBF}}(\mathbf{x},\mathbf{x}'\rvert\ell_{m})=\mathtt{\exp}\left(-\frac{\norm{\mathbf{x}-\mathbf{x}'}^{2}}{2\ell_{m}^{2}}\right).
\end{equation*}
Note that the length-scales $\{\ell_{m}\}_{m=1}^{M}$ are prefixed constants rather than trainable variables.
When applying the AK to a GPR, we optimize all the hyper-parameters $\{\alpha,\bm{\theta,\phi},\sigma\}$ by maximizing the marginal likelihood and make prediction as in GPR.

At first glance, the AK looks like a heuristic composite kernel.
However, the following sections explain how we design this kernel from the first principles.
In short, the kernel is distilled from a generative model called AKGPR that can naturally model non-stationary processes.

\subsection{A Generative Derivation of AK}\label{sec:two_type_nonstat}
\begin{figure}[tbp]%
    \centering
    \subfloat{%
        \resizebox{\linewidth}{!}{%
            \includegraphics[width=\linewidth,trim={5 5 5 5},clip]{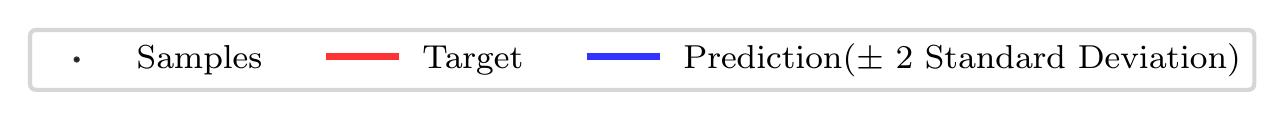}%
        }%
    }\vspace{-10pt}%
    \addtocounter{subfigure}{-1}\\%
    \subfloat[Wiggly Prediction\label{fig:1d_rbf_small_ell}]{%
        \resizebox{\linewidth}{!}{%
            \begin{tikzpicture}%
                \node {\includegraphics[width=\linewidth,trim={7 7 7 7},clip]{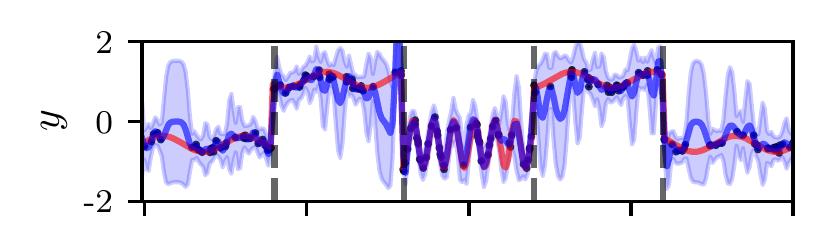}};%
                \node at (-1.95, 0.6) {\circled{1}};%
                \node at (-1.0, -0.5) {\circled{2}};%
                \node at (0.8, 0.6) {\circled{3}};%
                \node at (2.4, -0.5) {\circled{4}};%
                \node at (3.9, 0.6) {\circled{5}};%
            \end{tikzpicture}%
        }%
    }\vspace{-10pt}\\%
    \subfloat[Over-Smoothed Prediction\label{fig:1d_rbf_large_ell}]{%
        \resizebox{0.97\linewidth}{!}{%
    \includegraphics[width=\linewidth,trim={7 7 7 7},clip]{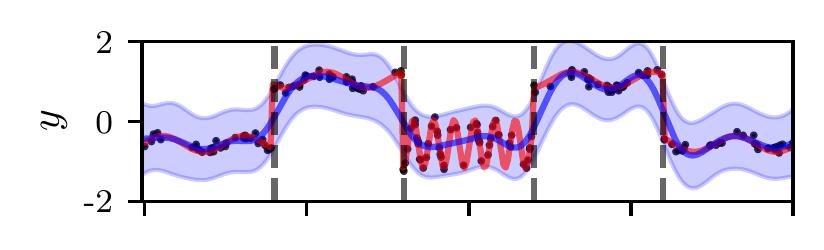}}}%
    \caption{\textbf{Learning A Non-Stationary Function using GPR with RBF Kernel}. The target function in red color consists of five partitions separated by vertical dashed lines. The black dots around the function are data points. The function changes drastically in partition\#3 and smoothly in the remaining partitions. The transitions between neighboring partitions are sharp. This simple function is challenging for a stationary kernel with a \textit{single} length-scale. GPR with a stationary RBF kernel produces either the wiggly prediction shown in \textbf{(a)} or the over-smoothed prediction in \textbf{(b)}. Note that, in \textbf{(a)}, the prediction in the smooth regions is rugged, and the uncertainty is over-conservative when the training data is sparse. The prediction in \textbf{(b)} only captures the general trend, and every input location seems equally uncertain.}\label{fig:1d_rbf}%
\end{figure}

The example in \Cref{fig:1d_rbf} motivates us to consider using different length-scales at different input locations.
Ideally, we need a smaller length-scale for partition\#3 and larger length-scales for the others.
In addition, we need to break the correlations among data points in different partitions.
An ideal non-stationary model should handle these two types of non-stationarity.
Many existing works model the input-dependent length-scale as a length-scale function~\citep{lang2007adaptive,plagemann2008nonstationary,heinonen2016nonstationary}.
However, parameter optimization of such models is sensitive to data distribution and parameter initialization.
We propose a new approach to address this issue that \emph{avoids learning an explicit length-scale function}.
Instead, every input location can \emph{select} among a set of GPs with different predefined primitive length-scales and \emph{select} which training samples are used when making a prediction.
This idea -- selecting instead of inferring an input-independent length-scale -- avoids optimization difficulties in prior work.
These ideas are developed in the following sections.

\begin{figure}[tbp]
    \centering%
    \subfloat{%
        \resizebox{\linewidth}{!}{
            \begin{tikzpicture}%
                \node at(0.0, 0.0){\includegraphics[width=\linewidth,trim={7 7 7 7},clip]{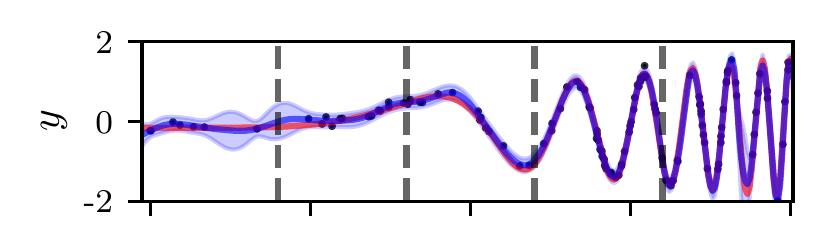}};%
                \node at(0.1, -2.4){\includegraphics[width=0.98\linewidth,trim={7 7 7 7},clip]{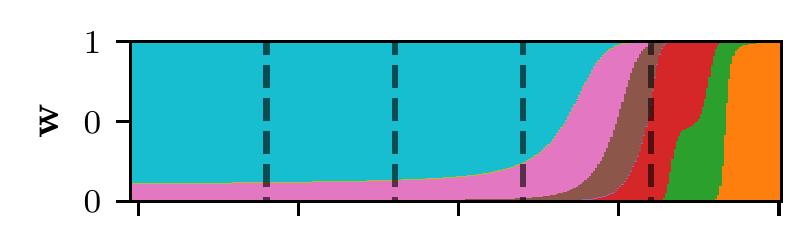}};%
                \node at(0.7, -3.9){\includegraphics[width=0.85\linewidth,trim={10 10 10 10},clip]{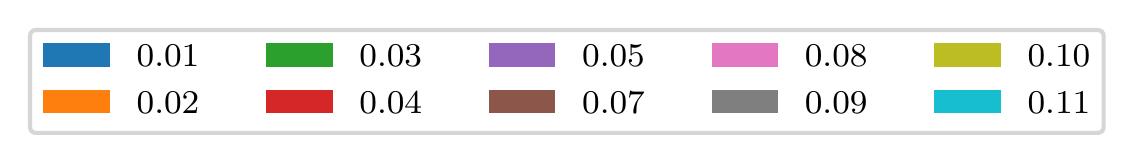}};%
            \end{tikzpicture}%
        }%
    }%
    \caption{\textbf{Learning $f(x)=x\sin(40x^{4})$ with Soft Length-Scale Selection}. The $\mathbf{w}$-plot visualizes the associated weighting vector $\mathbf{w}_{\bm{\theta}}(\mathbf{x})$ of each input location. The more vertical length a color occupies, the higher weight we assign to the GP with the corresponding length-scale. The set of predefined length-scales is color-labeled at the bottom. The learned weighting function gradually shifts its weight from smooth GPs to bumpy ones.}\label{fig:sin_weight}%
\end{figure}

\subsubsection{Length-Scale Selection}\label{sec:ls_selection}
Consider a set of $M$ independent GPs with a set of base kernels $\mathtt{k}_{m}(\mathbf{x},\mathbf{x}')$ using predefined primitive length-scales $\{\ell_{m}\}_{m=1}^{M}$.
Intuitively, if every input location can select a GP with an appropriate length-scale, the non-stationarity can be characterized well.
We can achieve this by an \emph{input-dependent weighted sum}
\begin{align}
  \mathtt{f}(\mathbf{x})&=\sum_{m}^{M}\mathtt{w}_{m}(\mathbf{x})\mathtt{g}_{m}(\mathbf{x}),\text{ where}\label{eq:f_gp}\\
  \mathtt{g}_{m}(\mathbf{x})&\sim\mathcal{GP}(\mathtt{0},\mathtt{k}_{m}(\mathbf{x},\mathbf{x}')).\label{eq:g_gp}
\end{align}
Here, $\mathtt{w}_{m}(\mathbf{x})$ is the $m$-th output of a vector-valued weighting function $\mathbf{w}_{\bm{\theta}}(\mathbf{x})$ which is parameterized by $\bm{\theta}$.
We denote $\mathbf{w}=[\mathtt{w}_{1}(\mathbf{x}),\dots,\mathtt{w}_{M}(\mathbf{x})]^{\T}$.

Consider an extreme case where $\mathbf{w}$ is a ``one-hot'' vector -- a binary vector with only one element being one and all other elements being zeros.
In this case, $\mathbf{w}$ selects a single appropriate GP depending on the input location.
Typically, inference techniques such as Gibbs sampling or Expectation Maximization are required for learning such discrete ``assignment'' parameters.
We lift this requirement by continuous relaxation:
\begin{equation}
  \mathbf{w}_{\bm{\theta}}(\mathbf{x})=\mathtt{\mathbf{softmax}}(\tilde{\mathbf{w}}_{\bm{\theta}}(\mathbf{x})),\label{eq:w_softmax}
\end{equation}
where $\tilde{\mathbf{w}}_{\bm{\theta}}(\mathbf{x})$ is an arbitrary $M$-dimensional function parameterized by $\bm{\theta}$.
Moreover, using such continuous weights has an advantage in modeling gradually changing non-stationarity, as shown in \Cref{fig:sin_weight}.

\Cref{fig:1d_weight} shows that length-scale selection gives better prediction after learning from the same dataset as in \Cref{fig:1d_rbf}.
However, when facing abrupt changes, as shown in the circled area, the model can only select a very small length-scale to accommodate the loose correlations among data.
If samples near the abrupt changes are not dense enough, a small length-scale might result in a high prediction error.
The following section explains how to handle abrupt changes using instance selection.

\subsubsection{Instance Selection}\label{sec:instance_selection}
Intuitively, an input-dependent length-scale specifies each data point's neighborhood radius that it can impact.
Simply varying the radius cannot handle abrupt changes, for example, in a step function, because data sampled before and after an abrupt change should break their correlations even when they are close in input locations.
We need to control the \emph{visibility} among samples: each sample learns only from other samples in the same subgroup.
To this end, we associate each input with a \emph{membership vector} $\mathbf{z}\triangleq\mathbf{z}_{\bm{\phi}}(\mathbf{x})$ and use a dot product between two membership vectors to control visibility.
Two inputs are visible to each other when they hold similar memberships.
Otherwise, their correlation will be masked out:
\begin{equation}\label{eq:instance_selection}
  \mathtt{k}_{m}(\mathbf{x},\mathbf{x}')=\mathbf{z}^{\T}\mathbf{z}'\mathtt{k}_{\text{RBF}}(\mathbf{x},\mathbf{x}'\rvert\ell_{m}).
\end{equation}
We can view this operation as input {\em dimension augmentation} where we append $\mathbf{z}$ to $\mathbf{x}$ but use a structured kernel in the joint space of $[\mathbf{x},\mathbf{z}]$.

Discussing one-hot vectors also helps understand the effect of $\mathbf{z}$.
In this case, the dot product is equal to $1$  if and only if $\mathbf{z}$ and $\mathbf{z}'$ are the same one-hot vector.
Otherwise, the dot product in \Cref{eq:instance_selection} masks out the correlation.
This way, we only use the subset of data points in the same group.
To make the model more flexible and simplify the parameter optimization, we again use soft memberships:
\begin{equation}\label{eq:z_softmax}
  \mathbf{z}_{\bm{\phi}}(\mathbf{x})=\mathtt{\mathbf{softmax}}(\tilde{\mathbf{z}}_{\bm{\phi}}(\mathbf{x})).
\end{equation}
Here, $\tilde{\mathbf{z}}_{\bm{\phi}}(\mathbf{x})$ is an arbitrary $M$-dimensional function parameterized by $\bm{\phi}$.

\input{figure_6_1d_weight.tex}

\subsubsection{The AKGPR Model}\label{sec:akgpr}
\begin{figure}[tbp]
    \centering
    \resizebox{0.8\linewidth}{!}{
        \begin{tikzpicture}
            \node[const] (pz) {$\bm{\phi}$};
            \node[latent, right=of pz] (z) {$\mathbf{z}_{n}$};
            \node[latent, right=of z] (g) {$g_{n,m}$};
            \node[const, right=of g] (l) {$\ell_{m}$};
            \plate {pm} {(g)(l)} {$M$};
        \node[obs, above=of g] (x) {$\mathbf{x}_{n}$};
        \node[latent, below=of l] (f) {$f_{n}$};
        \node[const, below=of pz] (a) {$\alpha$};
        \node[obs, right=of f] (y) {$y_{n}$};
        \node[latent, right=of l] (w) {$\mathbf{w}_{n}$};
        \node[const, right=of y] (s) {$\sigma$};
        \node[const, right=of w] (pw) {$\bm{\theta}$};
        \plate {pn} {(x)(z)(w)(g)(f)(y)} {$N$};
    \edge {x, pz} {z};
\edge {x, pw} {w};
\edge {x, z, l, a} {g};
\edge {w, g} {f};
\edge {f, s} {y};
\end{tikzpicture}
}
\caption{\textbf{Plate Notation of AKGPR}.}\label{fig:plate}%
\end{figure}
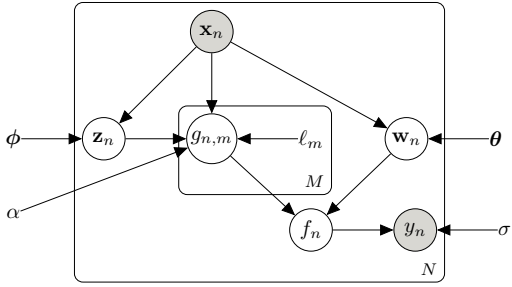

Combining the two ideas, we get a new probabilistic generative model developed for non-stationary environments called Attentive Kernel Gaussian Process Regression~(AKGPR).
Given $N$ inputs $\mathbf{X}\in\real^{N\times{D}}$ and targets $\mathbf{y}\in\real^{N}$, the model describes the generative process as follows.
We use some shorthands for compact notation: $\mathbf{g}_{m}\triangleq[\mathtt{g}_{m}(\mathbf{x}_{1}),\dots,\mathtt{g}_{m}(\mathbf{x}_{N})]^{\T},\mathbf{f}\triangleq[\mathtt{f}(\mathbf{x}_{1}),\dots,\mathtt{f}(\mathbf{x}_{N})]^{\T},\mathbf{w}_{m}\triangleq[\mathtt{w}_{m}(\mathbf{x}_{1}),\dots,\mathtt{w}_{m}(\mathbf{x}_{N})]^{\T}$.
Here $\mathtt{w}_{m}(\mathbf{x})$ is the $m$-output of \Cref{eq:w_softmax}.
\begin{itemize}
  \item We compute the membership vector $\mathbf{z}_{n}$ for each input using \Cref{eq:z_softmax}. Plugging $\mathbf{z}_{n}$ and the predefined length-scales $\ell_{m}$ into \Cref{eq:instance_selection}, we then compute $M$ covariance matrices $\mathbf{K}_{m}$ evaluated at every pair of inputs.
  \item The vector $\mathbf{g}_{m}$ follows a multivariate Gaussian distribution $\mathcal{N}(\mathbf{0},\mathbf{K}_{m})$ according to the definition of GPs and \Cref{eq:g_gp}. From \Cref{eq:f_gp}, we can see that $\mathbf{f}$ is the summation of $M$ vectors that follows affine-transformed multivariate Gaussian distributions, thus $\mathbf{f}$ also follows Gaussian distribution:
    \begin{equation}\label{eq:akgpr_f}
      \mathbf{f}=\sum_{m=1}^{M}\mathbf{W}_{m}\mathbf{g}_{m}\sim\mathcal{N}(\mathbf{0},\sum_{m=1}^{M}\mathbf{W}_{m}\mathbf{K}_{m}\mathbf{W}_{m}^{\T}),
    \end{equation}
    where $\mathbf{W}_{m}$ is a diagonal matrix with $\mathbf{w}_{m}$ being the $N$ diagonal elements.
  \item Finally, we can generate the targets $\mathbf{y}$ using the Gaussian likelihood in \Cref{eq:likelihood}.
\end{itemize}
The plate diagram of this generative process is shown in \Cref{fig:plate}.
From \Cref{eq:akgpr_f} we observe that the generative process of AKGPR is equivalent to that of a GPR with a new kernel:
\begin{equation}\label{eq:raw_ak}
  \mathtt{k}(\mathbf{x},\mathbf{x}')=\sum_{m=1}^{M}w_{m}\underbracket{\mathbf{z}^{\T}\mathbf{z}'\mathtt{k}_{\text{RBF}}(\mathbf{x},\mathbf{x}'\rvert\ell_{m})}_{\text{hidden in }\mathbf{K}_{m}}w_{m}'.
\end{equation}
Since $\mathbf{z}^{\T}\mathbf{z}'$ is independent of $m$, we can move it outside the summation to avoid duplicate computation.

\Cref{eq:raw_ak} is almost the same as the AK in \Cref{def:attentive_kernel}, except that this kernel is not normalized yet.
When $\mathbf{x}=\mathbf{x}'$, the kernel value $\mathtt{k}(\mathbf{x},\mathbf{x}')$ might be greater than $1$.
As mentioned in \Cref{sec:model_specification}, using an amplitude parameter $\alpha$ to adjust the scale of the kernel value is a common practice in GPR.
Introducing the amplitude hyper-parameter requires the kernel to be normalized; otherwise, the interplay between the amplitude and the scaling effect of a kernel before normalization makes the optimization difficult because more local optima are introduced due to the symmetries of the parameter space.
We normalize $\mathbf{w}$ and $\mathbf{z}$ with $\ell^{2}$\,-\,norm to ensure that the maximum kernel value (when $\mathbf{x}=\mathbf{x}'$) is $1$, and $\alpha$ is the only parameter that controls the scale of kernel value.
After normalization, we now have the final version of the proposed AK in \Cref{def:attentive_kernel}, which can be used in any GP model.
From the discussion above we have:
\begin{proposition}\label{prop:equivalence}
  The AKGPR generative model is equivalent to a GPR model with the AK defined in~\Cref{def:attentive_kernel}.
\end{proposition}

\input{figure_8_1d_ak.tex}

\subsection{Applying AK to GPR}
We use the AK with a GPR model and optimize all the parameters by maximizing the log marginal likelihood $\ln{p(\mathbf{y}\rvert{\sigma,\alpha,\bm{\theta,\phi}})}$.
\Cref{fig:1d_ak} shows the prediction results on the example from \Cref{fig:1d_rbf}.
Now we can accurately model the highly varying part, the smooth parts, and the abrupt changes.
Compared to \Cref{fig:1d_rbf}, where the uncertainty mainly depends on the proximity to training samples, the AKGPR assigns higher uncertainty to the high-error locations.
The better uncertainty quantification is achieved by putting more weight on the GPs with small length-scales in partition\#3 and those with large length-scales in the other partitions.
Note that the AKGPR switches the membership vector $\mathbf{z}$ in the circled area to mask the inter-partition correlations, which cannot be realized by length-scale selection in \Cref{fig:1d_weight}.
Due to this modeling advantage, the results in \Cref{fig:1d_ak} are qualitatively better than in \Cref{fig:1d_weight}.

\subsection{Remark on The Attentive Kernel}
In this section, we discuss how to parameterize the weighting and membership functions in the AK, the computational complexity of the proposed kernel, and some details on hyper-parameter optimization of non-stationary kernels.

\subsubsection{Parameterization}
To instantiate an AK, we must specify the weighting function $\mathbf{w}_{\bm{\theta}}(\mathbf{x})$ and the membership function $\mathbf{z}_{\bm{\phi}}(\mathbf{x})$.
In the experiments, we find that sharing a single neural network for length-scale selection and instance selection does not affect the performance but reduces the number of trainable parameters and sometimes helps the training of the instance selection mechanism (see \Cref{sec:ablation}).
Therefore, we use the same set of parameters $\bm{\theta}=\bm{\phi}$ for the two attention mechanisms and choose a simple neural network with two hidden layers (see \Cref{sec:models} for more details).
Using a simple neural network is an arbitrary choice for simplicity and modeling flexibility.
Other parametric functions can also be used, and we leave the study of alternative parameterization to future work.

\subsubsection{Computational Complexity}
Kernel matrix computations are typically performed in a batch manner to take advantage of the parallelism in linear algebra libraries.
\Cref{fig:ak_diagram} shows the computational diagram of the self-covariance matrix of an input matrix $\mathbf{X}\in\real^{N\times{D}}$ for the case where the same function parameterizes $\mathbf{\mathtt{w}}_{\theta}(\mathbf{x})$ and $\mathbf{\mathtt{z}}_{\phi}(\mathbf{x})$.
\input{figure_9_ak_diagram.tex}
The computation of a cross-covariance matrix and the case where $\mathbf{\mathtt{w}}_{\theta}(\mathbf{x})$ and $\mathbf{\mathtt{z}}_{\phi}(\mathbf{x})$ are parameterized separately are handled similarly.
We first pass $\mathbf{X}$ to a neural network with two hidden layers to get $\mathbf{W}\in\real^{N\times{M}}\text{ and }\mathbf{Z}\in\real^{N\times{M}}$.
The computational complexity of this step is $\mathcal{O}(NDH+NH^2+NHM)$.
Then, we compute a visibility masking matrix $\mathbf{O}=\mathbf{Z}\mathbf{Z}^{\T}$, which takes $\mathcal{O}(N^{2}M)$.
After getting the pairwise distance matrix $\left(\mathcal{O}(N^{2}D)\right)$, we can compute the base kernel matrices using different length-scales $\left(\mathcal{O}(N^{2})\right)$.
The $m$-th kernel matrix is scaled by the outer-product matrix of the $m$-th column of $\mathbf{W}$, which takes $\mathcal{O}(N^{2}M)$.
Finally, we sum up the scaled kernel matrices and multiply the result with the visibility masking matrix to get the AK matrix $\left(\mathcal{O}(N^{2}M)\right)$.
We defer the discussion of the choices of network size $H$ and number of base kernels $M$ to the sensitivity analysis in \Cref{sec:sensitivity}.
In short, these will be relatively small numbers, so the overall computational complexity is still $\mathcal{O}(N^{2}D)$.
In practice, we find that the runtime of the AK experiments is around three times slower than that of the RBF kernel.

\subsubsection{Optimization}
We note that the model complexity term discussed in \Cref{sec:learning} is insufficient for preventing over-fitting when training non-stationary kernels for many iterations, a point also mentioned in~\cite{tompkins2020sparse} in their over-fitting analysis.
Although the AK is more robust to over-fitting (see \Cref{sec:overfitting}), we implement an incremental training scheme to improve the computational efficiency and optimization robustness when using non-stationary kernels in RIG.
Specifically, we train the model on all the collected data for $N_{t}$ iterations after collecting $N_{t}$ samples at the $t$-th decision epoch, which corresponds to line\,7 to line\,10 in \Cref{alg:rig_ak}.

This training scheme can be considered a rule-of-thumb early-stopping regularization.
We also find that, when using a neural network in a non-stationary kernel, the initial learning rate of the network parameters should be smaller than that of other hyper-parameters.
For example, when using the AK, the initial learning rates of $\bm{\theta}$ or $\bm{\phi}$ should be smaller than that of $\{\alpha,\sigma\}$.

Another important aspect is when to start optimizing the hyper-parameters.
Optimizing the parameters when the data is too sparse and not representative can lead to wrong length-scale prediction, which can bias the informative planning.
In RIG, exploring the environment and sampling data at different locations is necessary before optimizing the hyper-parameters.
In our experiments, this is done by following a predefined B\'ezier curve that explores the environment.
An alternative way to achieve this behavior is by fixing the hyper-parameters to some appropriate values and training the model only after collecting a certain amount of samples.
This approach does not require a pilot survey of the environment.
However, the user should have some prior knowledge of the target environment in order to set the initial hyper-parameters.

This training setup works well empirically, but we acknowledge that developing more principled ways to learn non-stationary GPs is an essential future direction, which is still an open research problem and has recently received increasing attention~\citep{ober2021promises,van2021feature,lotfi2022bayesian}.

{\algrenewcommand\textproc{}
  \begin{algorithm}[t!]
    \setstretch{1.1}\small%
    \caption{\textbf{Active Mapping with the AK}.}\label{alg:rig_ak}%
    \hspace*{\algorithmicindent} \textbf{Arguments}: $N_{\text{max}},\alpha,\sigma,\{\mathtt{k}_{m}(\mathbf{x},\mathbf{x}')\}_{m=1}^{M}$\\
    \hspace*{\algorithmicindent}\hspace{48pt} $\mathbf{\mathtt{w}}_{\bm{\theta}}(\mathbf{x}),\mathbf{\mathtt{z}}_{\bm{\phi}}(\mathbf{x}),\mathtt{strategy}$
    \begin{algorithmic}[1] 
      \State compute normalization and standardization statistics
      \State $\mathtt{kernel\leftarrow\mathtt{AK}(\alpha,\{\mathtt{k}_{m}(\mathbf{x},\mathbf{x}')\}_{m=1}^{M},\mathbf{\mathtt{w}}_{\bm{\theta}}(\mathbf{x}),\mathbf{\mathtt{z}}_{\bm{\phi}}(\mathbf{x}))}$
      \State $\mathtt{model\leftarrow\mathtt{GPR}(\mathtt{kernel},\sigma)}$
      \State $t\leftarrow{0}$
      \While{$\mathtt{model}.N_{\text{train}}<N_{\text{max}}$}\Comment{sampling budget}
      \State $\mathtt{\mathbf{x}_{\text{info}}\leftarrow\mathtt{strategy}(model)}$\Comment{informative waypoint}
      \State $\mathtt{\mathbf{X}_{t},\mathbf{y}_{t}\leftarrow{tracking\_and\_sampling}(\mathbf{x}_{\text{info}})}$\Comment{$N_{t}$ samples}
      \State $\mathtt{\bar{\mathbf{X}}_{t},\bar{\mathbf{y}}_{t}\leftarrow{normalize\_and\_standardize}(\mathbf{X}_{t},\mathbf{y}_{t}})$
      \State $\mathtt{model.add\_data(\bar{\mathbf{X}}_{t},\bar{\mathbf{y}}_{t}})$
      \State $\mathtt{model.optimize}(N_{t})$\Comment{maximize marginal likelihood}
      \State $t\leftarrow{t+1}$
      \EndWhile
      \State \textbf{return} $\mathtt{model}$
    \end{algorithmic}
  \end{algorithm}
}

\subsection{Active Mapping with The Attentive Kernel}
\Cref{alg:rig_ak} shows how the AK can be used for active mapping.
The system requires the following input arguments: the maximum number of training data $N_{\text{max}}$, the initial kernel amplitude $\alpha$, the initial noise scale $\sigma$, a set of $M$ base kernels $\{\mathtt{k}_{m}(\mathbf{x},\mathbf{x}')\}_{m=1}^{M}$, functions $\mathbf{\mathtt{w}}_{\bm{\theta}}(\mathbf{x})$, $\mathbf{\mathtt{z}}_{\bm{\phi}}(\mathbf{x})$, and a sampling strategy.
First, we need to compute the statistics to normalize the inputs $\mathbf{X}$ roughly to the range $[-1,1]$ and standardize the targets $\mathbf{y}$ to nearly have zero mean and unit variance~(line\,1).
We can get these statistics from prior knowledge of the environment.
The workspace extent is typically known, allowing the normalization statistics to be readily calculated.
The target-value statistics can be rough estimates or computed from a pilot environment survey~\citep{kemna2018pilot}.
Then, we instantiate an AK and a GPR with the given parameters~(lines\,2-3).
At each decision epoch $t$, the sampling strategy proposes informative waypoints by optimizing an objective function derived from the predictive uncertainty of the GPR~(line\,6).
The robot tracks the informative waypoints and collects $N_{t}$ samples along the trajectory~(line\,7).
Note that the number of collected samples is typically larger than the number of informative waypoints.
The new samples are normalized and standardized and then appended to the model's training set~(lines\,8-9).
Finally, we maximize the log marginal likelihood for $N_{t}$ iterations~(line\,10).
The robot repeats predicting~(hidden in line\,6), planning, sampling, and optimizing until the sampling budget is exceeded~(line\,5).

\begin{figure*}[tbp]
    \centering
    \setlength{\fboxrule}{5pt}
    \setlength{\fboxsep}{0pt}
    \subfloat[N17E073 3D Perspective\label{fig:N17E073}]{
        \resizebox{0.5\linewidth}{!}{
            \includegraphics[width=\linewidth]{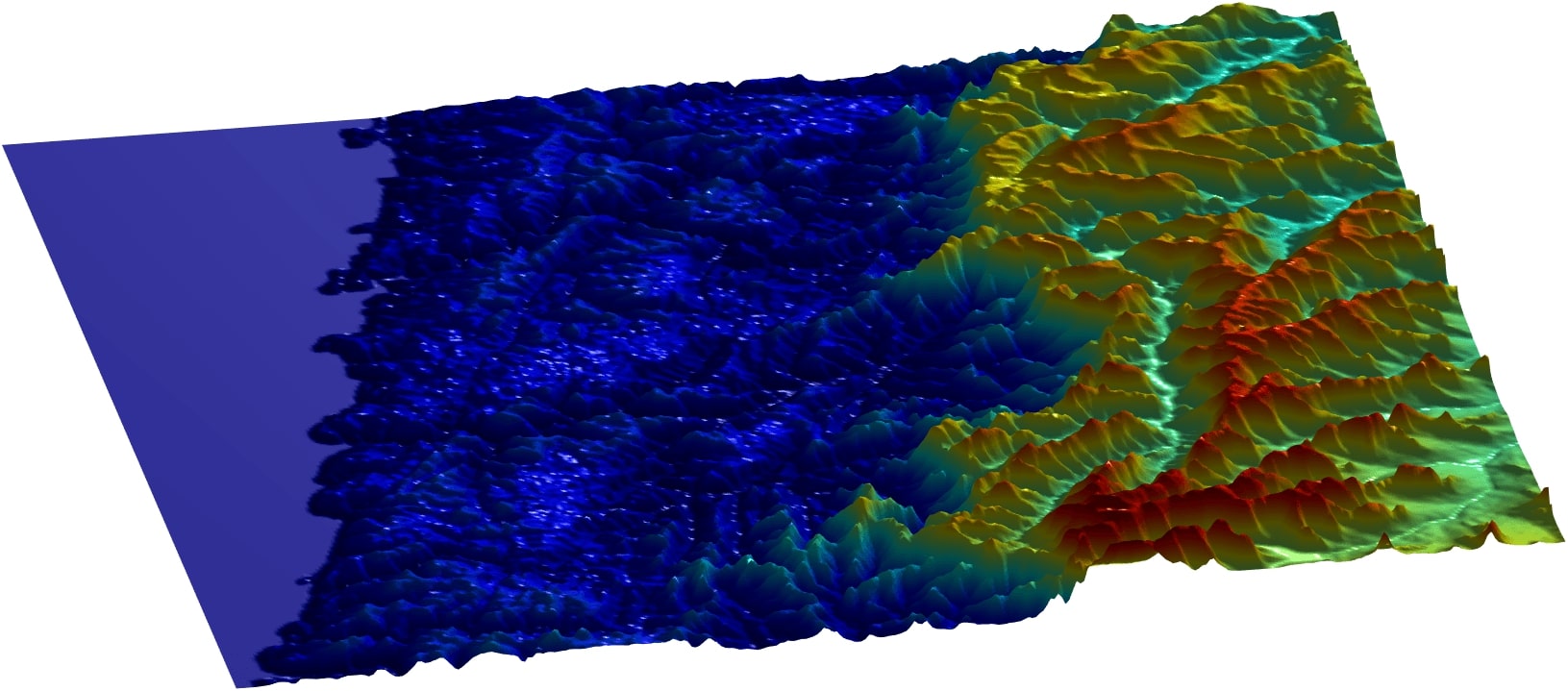}\hspace{1em}
        }
    }
    \subfloat[N17E073 Bird's-Eye View]{
        \resizebox{0.45\linewidth}{!}{
            \fbox{\includegraphics[width=\linewidth]{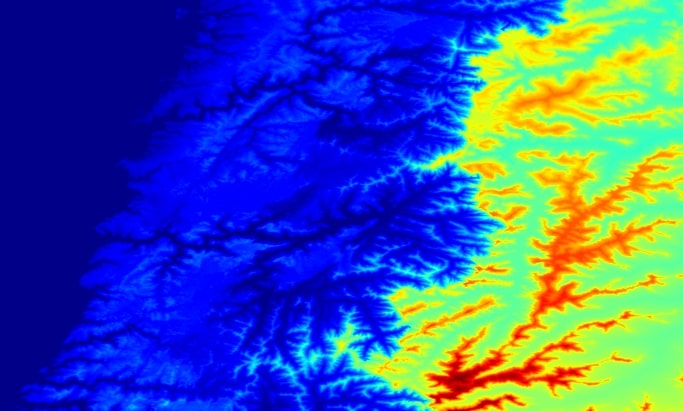}}
        }
    }\\
    \subfloat[N43W080 3D Perspective\label{fig:N43W080}]{
        \resizebox{0.5\linewidth}{!}{
            \begin{tikzpicture}
                \node(a){\includegraphics[width=\linewidth]{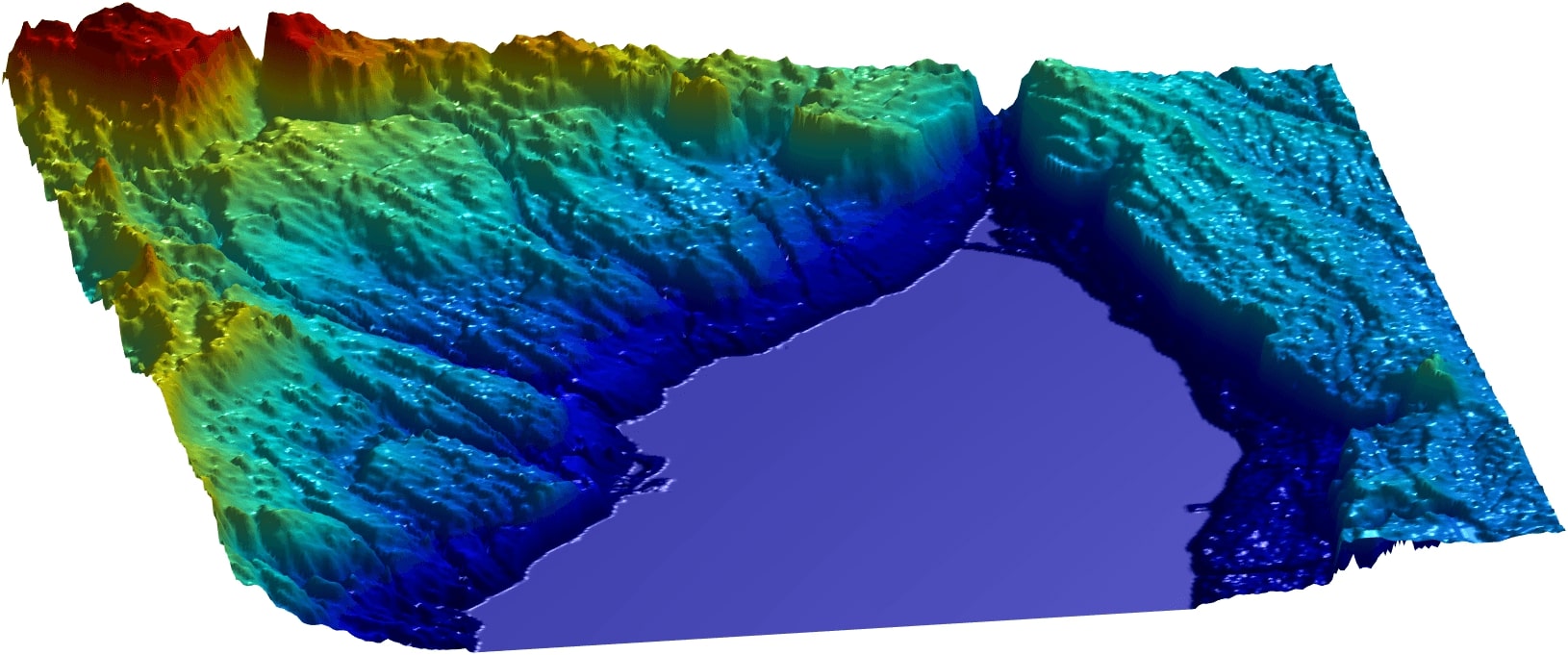}\hspace{1em}};
                \draw [orange,-stealth,line width=3](3,3.5) -- (2.5,3);
                \draw [orange,-stealth,line width=3](-6.5,3.8) -- (-5.8,3.3);
            \end{tikzpicture}
        }
    }
    \subfloat[N43W080 Bird's-Eye View]{
        \resizebox{0.45\linewidth}{!}{
            \fbox{\includegraphics[width=\linewidth]{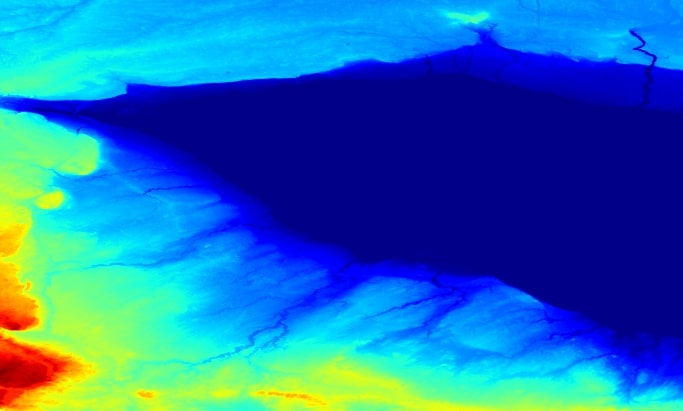}}
        }
    }\\
    \subfloat[N45W123 3D Perspective\label{fig:N45W123}]{
        \resizebox{0.5\linewidth}{!}{
            \includegraphics[width=\linewidth]{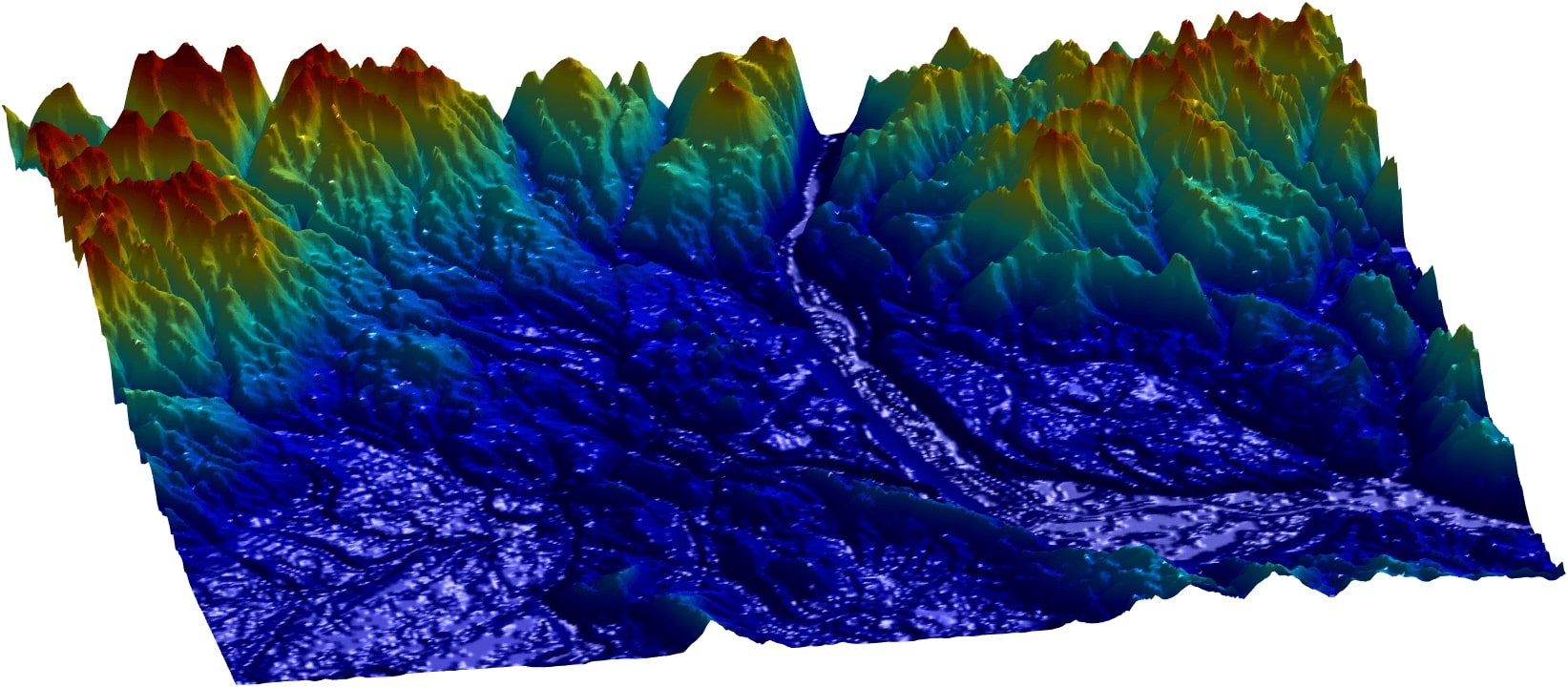}\hspace{1em}
        }
    }
    \subfloat[N45W123 Bird's-Eye View\label{fig:env_N45W123}]{
        \resizebox{0.45\linewidth}{!}{
            \fbox{\includegraphics[width=\linewidth]{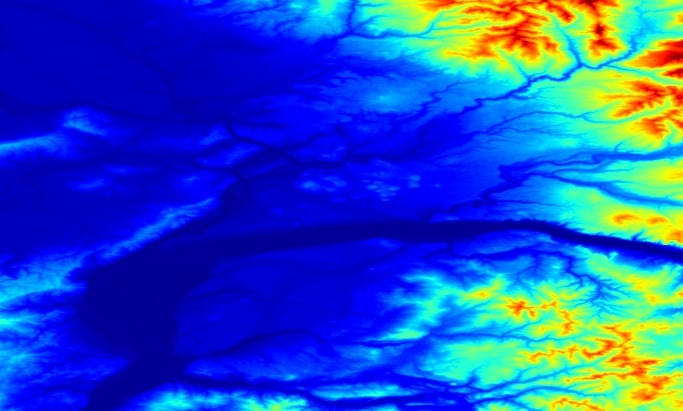}}
        }
    }\\
    \subfloat[N47W124 3D Perspective\label{fig:N47W124}]{
        \resizebox{0.5\linewidth}{!}{
            \includegraphics[width=\linewidth]{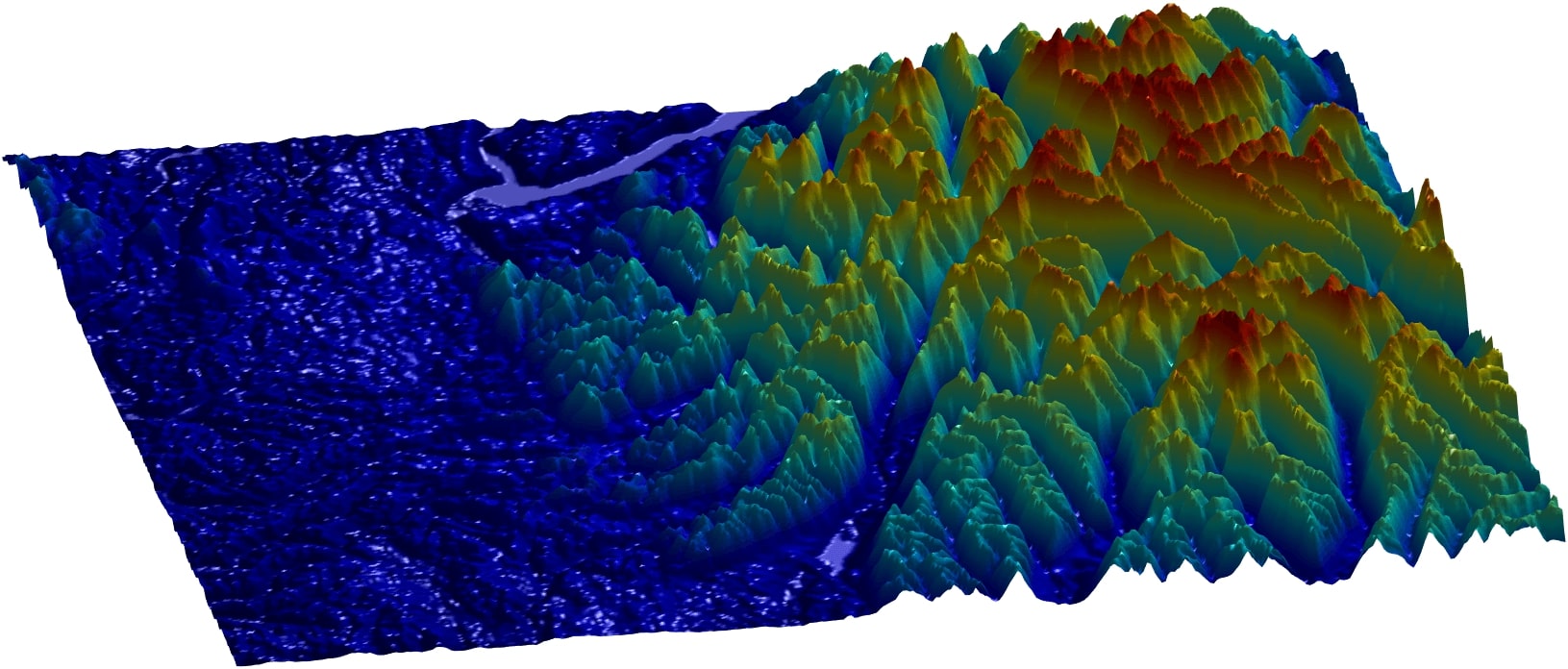}\hspace{1em}
        }
    }
    \subfloat[N47W124 Bird's-Eye View\label{fig:env_N47W124}]{
        \resizebox{0.45\linewidth}{!}{
            \fbox{\includegraphics[width=\linewidth]{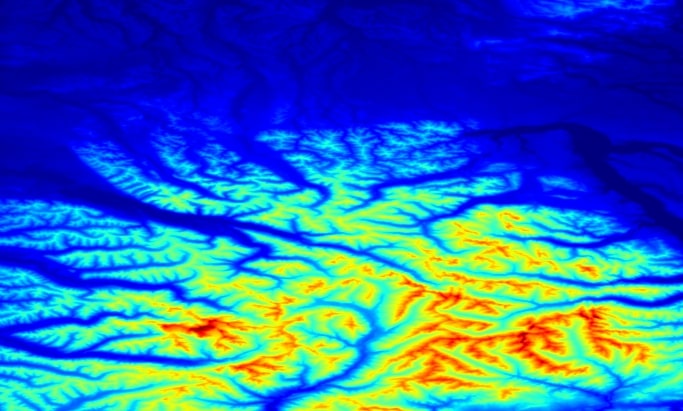}}
        }
    }
    \caption{\textbf{The Four Environments Used in the Elevation Mapping Tasks}. Note that the 3D perspectives are rotated and rescaled to highlight the visual features of the environments.}
    \label{fig:envs}
\end{figure*}

\begin{figure}[tbp]
    \centering
    \includegraphics[width=\linewidth]{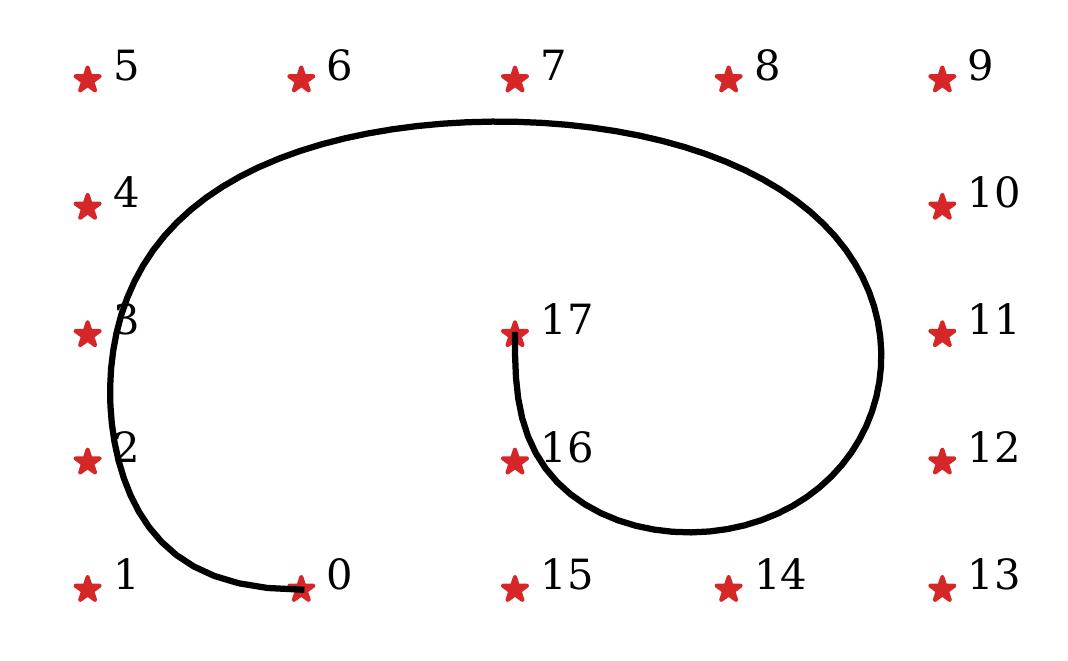}
    \caption{\textbf{Pilot Survey Path}. The red stars are control points to generate the B\'ezier curve.}\label{fig:bezier}%
\end{figure}

\section{Experiments}\label{sec:experiments}
We design our experiments to address the following questions.
\begin{itemize}
  \item[\textbf{Q1}] How does the AK compare to its stationary counterpart and other non-stationary kernels in prediction accuracy and uncertainty quantification?
  \item[\textbf{Q2}] If non-stationary kernels have better uncertainty quantification capability, can we use the uncertainty for active data collection and to further improve the prediction accuracy?
  \item[\textbf{Q3}] Some parameters in the AK need to be determined, \textit{i.e.}, the number and range of the primitive length-scales and the network hyper-parameters. Are these parameters hard to tune? Is the performance of AK sensitive to its parameter settings?
  \item[\textbf{Q4}] The AK consists of two ideas: length-scale selection and instance selection. Which one contributes more to the performance in the experiments?
  \item[\textbf{Q5}] How does the AK compare to the other non-stationary kernels in over-fitting?
\end{itemize}

To answer \textbf{Q1}, we use random sampling experiments in \Cref{sec:random_sampling} to evaluate the AK and the compared kernels.
We run the random sampling experiments first because the performance of a RIG system depends on not only the model's prediction and uncertainty but also the informative planner.
Sampling data uniformly at random (without an informative planner) provides controlled experiments to understand the effects of using different kernels.
For \textbf{Q2}, we conduct both active learning~(\Cref{sec:active_sampling}) and RIG experiments~(\Cref{sec:informative_planning}) to disentangle the influence of the model's uncertainty and the planner.
RIG considers the physical constraints of the robot embodiment, while active learning can sample arbitrary locations.
We assess the AK via sensitivity analysis, ablation study, and over-fitting analysis to address the remaining questions.

\subsection{Simulated Experiments}\label{sec:simulated_experiments}
We have conducted extensive simulations in four representative environments that exhibit various non-stationary features.
The elevation maps are downloaded from the NASA Shuttle Radar Topography Mission (\href{https://dwtkns.com/srtm30m/}{dwtkns.com/srtm30m}).
Supplemental materials can be found at \href{https://weizhe-chen.github.io/attentive_kernels/}{weizhe-chen.github.io/attentive\_kernels}.

\subsubsection{Environments}\label{sec:environments}
\Cref{fig:envs} shows the 3D perspectives of all the environments and the corresponding bird's-eye views.
Note that the 3D plots are rotated for better visualization.
When comparing to the model prediction, we use the bird's-eye map as the ground truth, and we will describe the environmental features in the 3D plots.
Looking at environment \texttt{N17E073} from left to right, it consists of a flat part, a mountainous area, and a rocky region with many ridges.
A good non-stationary GP model should use decreasing length-scales from left to right.
Also, note that the most complex area (\textit{i.e.}, the red region) occupies roughly one-third of the whole environment.
\texttt{N43W080} presents sharp elevation changes indicated by the arrows while the lakebed is virtually flat.
Using a large length-scale can model most of the areas well, albeit better prediction can be achieved by sampling densely around the high-variability spots indicated by the arrows.
It is worth noting that better predictions will be more evident in the visualization compared to the evaluation metrics that average across the whole environment since the important area only occupies a small portion of the environment, and the improvements might be negligible in the metrics.
In \texttt{N45W123}, the environment has a narrow complex upper part and a smoother lower part.
The size of the complex region is smaller than one-third of the environment.
There is also a ``river'' passing through the middle.
The right part of \texttt{N47W124} varies drastically, while its left part is relatively flat.
Loosely speaking, \texttt{N47W124} has the most significant change in spatial variability, followed by \texttt{N17E073} and then \texttt{N45W123}, so the possible improvement margins of non-stationary models in these environments should also decrease in this order.
Only after discovering and sampling the two arrow-indicated spots can non-stationary models show an advantage over a stationary one in predicting environment \texttt{N43W080}.

\subsubsection{Robot}\label{sec:sim_robot}
We set the extent of the environment to $20\times{20}$ meters and simulate a planar robot that has a simple Dubins' car model $[\dot{x}_{1},\dot{x}_{2},\dot{v},\dot{\omega}]=[v\cos(\omega),v\sin(\omega),a_{1}, a_{2}]$.
Here, $\mathbf{x}_{b}\triangleq[x_{1}, x_{2}]^{\T}$ is the position, $\omega\in[-\pi,\pi)$ is the orientation, and $\mathbf{a}\triangleq[a_{1},a_{2}]^{\T}$ is the action that represents the change in the linear velocity and angular velocity.
The maximum linear velocity is set to $1\ m/s$, and the control frequency is $10\ Hz$.
Although we assume perfect localization in the simulated experiments, to keep the same interface with the field experiments, we consider that the robot has achieved a goal if it is within a $0.1$-meter radius.
This radius is an arbitrary choice within the dimension of the robot.
The robot has a single-beam range sensor that collects one noisy elevation measurement per second with a unit Gaussian observational noise.
In the random and active sampling experiments, the robot can ``jump'' to an arbitrary sampling location to collect data, so it does not follow Dubins' car model.
In the RIG experiment, the robot tracks some informative locations under the Dubins' car kinematic constraint and collects elevation measurements along its trajectory.

\subsubsection{Models}\label{sec:models}
The GPR takes two-dimensional sampling locations as inputs and predicts the elevation.
We only allow the robot to collect $700$ samples, among which the first $50$ data points are collected along a pilot survey path pre-computed for the environment.
As shown in \Cref{fig:bezier}, the path is generated from a B\'ezier curve with $18$ control points. The positions of the control points adapt to the extent of the workspace accordingly.
These $50$ samples are used to initialize the GPR and compute the statistics to normalize the inputs and standardize the target values.
If the statistics are known in advanced, the pilot survey is not necessary. One can also use a relatively large length-scale and fix the hyper-parameters of the GPR in the early stage so that the robot can explore the environment and collect diverse data for hyper-parameter optimization and statistics calculation.
After normalization and standardization, we initialize the hyper-parameters to $\ell=0.5,\alpha=1.0,\sigma=1.0,\ell_{\text{min}}=0.01,\ell_{\text{max}}=0.5$.
We use the default PyTorch settings for initializing the network parameters.
These hyper-parameters and the neural network parameters in the non-stationary kernels are jointly optimized by two Adam optimizers~\citep{kingma2014adam} with initial learning rates $0.01$ and $0.001$, respectively.
We first run an initial optimization of all the parameters for $50$ steps.
The model's prediction is evaluated on a $50\times{50}$ linearly spaced evaluation grid, \textit{i.e.}, $2500$ query inputs, comparing with the ground-truth elevation values.

We compare the AK with two existing non-stationary kernels: the Gibbs kernel and Deep Kernel Learning (DKL).
Since the RBF kernel is widely used in RIG, we also add this kernel as a stationary baseline.
The Gibbs kernel extends the length-scale to be any positive function of the input, degenerating to an RBF kernel when using a constant length-scale function.
Following \cite{remes2018neural}, which showed improved results, the length-scale function is modeled using a neural network instead of another Gaussian process.
DKL addresses non-stationarity through input warping.
A neural network transforms the inputs to a feature space where the stationary RBF kernel is assumed to be sufficient.
We use the same neural network with $2\times{10}\times{10}\times{10}$ neurons and hyperbolic tangent activation function for the AK and DKL and change the output dimension to $1$ for the Gibbs kernel because it requires a scalar-valued length-scale function.

{\algrenewcommand\textproc{}
  \begin{algorithm}[t!]
    \setstretch{1.1}\small%
    \caption{\textbf{A Myopic Informative Planning Strategy}.}\label{alg:rig_strategy}%
    \hspace*{\algorithmicindent} \textbf{Notation}\\
    \hspace*{\algorithmicindent} $\mathtt{workspace}$: bounding box of the workspace\\
    \hspace*{\algorithmicindent} $\mathtt{N_{c}}$: number of candidate locations\\
    \hspace*{\algorithmicindent} $\mathtt{model}$: Gaussian process regression model\\
    \hspace*{\algorithmicindent} $\mathtt{\mathbf{x}_{b}}$: robot's position
    \begin{algorithmic}[1] 
      \Procedure{$\mathtt{rig}$}{$\mathtt{workspace,N_{c},model,pose}$}
      \State $\mathtt{\mathbf{X}_{c}\sim\mathcal{U}(workspace,N_{c})}$\Comment{generate candidate locations uniformly at random in the workspace}
      \State $\bm{\mu,\nu}\leftarrow\mathtt{model.predict(\mathbf{X}_{c})}$\Comment{predictive mean and variance}
      \State $\bm{\varepsilon}\leftarrow\mathtt{entropy}(\bm{\nu})$\Comment{compute predictive entropy}
      \State $\mathbf{d}\leftarrow\mathtt{distance(\mathbf{X}_{c},\mathbf{x}_{b})}$\Comment{pairwise Euclidean distances}
    \State $\bar{\bm{\varepsilon}}\leftarrow\nicefrac{\left[\bm{\varepsilon}-\mathtt{min}(\bm{\varepsilon})\right]}{\left[\mathtt{max}(\bm{\varepsilon})-\mathtt{min}(\bm{\varepsilon})\right]}$
  \State $\bar{\bm{d}}\leftarrow\nicefrac{\left[\bm{d}-\mathtt{min}(\bm{d})\right]}{\left[\mathtt{max}(\bm{d})-\mathtt{min}(\bm{d})\right]}$
      \State $\mathtt{score}\leftarrow\bm{\varepsilon}-\mathbf{d}$\Comment{informativeness score}
      \State $i\leftarrow\argmax(\mathtt{score})$
      \State \textbf{return} the $i$-th candidate in $\mathbf{X}_{c}$
      \EndProcedure
    \end{algorithmic}
  \end{algorithm}
}

\subsubsection{Sampling Strategies}\label{sec:sampling_strategies}
We use different sampling strategies in the three sets of experiments.
We randomly draw a sample from a uniform distribution at each decision epoch in random sampling experiments.
In active sampling experiments, we evaluate the predictive uncertainty on $1000$ randomly generated candidate locations and then sample from the location with the highest predictive entropy.
While the AK can be plugged into any advanced informative planner for RIG, we use the naive informative planner in \Cref{alg:rig_strategy} for simplicity.
Specifically, in addition to the predictive entropy, this planner computes the distances from these locations to the robot's position.
We normalize the predictive entropy and distance to $[0, 1]$.
Each candidate location's informativeness score is defined as the normalized entropy minus the normalized distance.
This informativeness score considers the robot's physical constraints and encourages the robot to move to a location with high predictive uncertainty and close to the robot's current position.
The planner outputs the informative waypoint with the highest score.
A tracking controller is used to move the robot to the waypoint.
Note that the number of collected samples $N_{t}$ varies at different decision epochs depending on the distance from the robot to the informative waypoint.

\rowcolors{2}{gray!25}{white}
\begin{table*}[tbp]
  \centering\small
  \caption{\textbf{Random Sampling Benchmarking Results}.}\label{tab:random}
  \begin{tabular}{
      ll
      S[table-format=1.2(3)e-1]
      S[table-format=-1.2(3)e-1]
      S[table-format=1.2(3)]
      S[table-format=1.2(3)e1]
      S[table-format=1.2(3)e1]
    }
    \toprule
    {\textbf{Environment}} & {\textbf{Method}} & {\textbf{SMSE}$\downarrow^{1}_{0}$} & {\textbf{MSLL}$\downarrow^{0}$} & {\textbf{NLPD}$\downarrow$} & {\textbf{RMSE}$\downarrow_{0}$} & {\textbf{MAE}$\downarrow_{0}$} \\
    \midrule
    N17E073   & RBF   &            1.33(3)e-01 &             -9.9(1)e-01 &            4.59(1)e+00 &            2.33(3)e+01 &            1.69(3)e+01 \\
              & AK    &  \B 1.11(4)e-01 &  \B -1.24(1)e+00 &  \B 4.34(1)e+00 &  \B 2.13(4)e+01 &  \B 1.50(2)e+01 \\
              & Gibbs &            1.33(1)e-01 &            -1.09(2)e+00 &            4.50(3)e+00 &            2.33(9)e+01 &            1.66(4)e+01 \\
              & DKL   &            1.37(6)e-01 &             -9.7(3)e-01 &            4.62(3)e+00 &            2.37(5)e+01 &            1.68(4)e+01 \\
    N43W080   & RBF   &            7.1(3)e-02 &            -1.43(2)e+00 &            3.87(2)e+00 &            1.23(3)e+01 &           8.13(6)e+00 \\
              & AK    &  \B 6.0(5)e-02 &  \B -1.69(6)e+00 &  \B 3.62(6)e+00 &  \B 1.11(5)e+01 &  \B 7.0(2)e+00 \\
              & Gibbs &            7.2(4)e-02 &            -1.48(6)e+00 &            3.83(6)e+00 &            1.25(5)e+01 &            8.3(3)e+00 \\
              & DKL   &            6.6(8)e-02 &            -1.49(4)e+00 &            3.81(4)e+00 &            1.19(7)e+01 &            7.5(3)e+00 \\
    N45W123   & RBF   &            1.65(7)e-01 &             -9.4(3)e-01 &            4.37(3)e+00 &            1.97(4)e+01 &            1.28(3)e+01 \\
              & AK    &  \B 1.41(6)e-01 &  \B -1.28(2)e+00 &  \B 4.03(2)e+00 &  \B 1.80(4)e+01 &  \B 1.15(2)e+01 \\
              & Gibbs &             1.8(1)e-01 &            -1.08(1)e+00 &            4.24(2)e+00 &            2.07(7)e+01 &            1.34(2)e+01 \\
              & DKL   &             2.0(1)e-01 &             -9.1(1)e-01 &            4.41(1)e+00 &            2.18(7)e+01 &            1.42(6)e+01 \\
    N47W124   & RBF   &            2.26(7)e-01 &             -7.2(1)e-01 &            4.77(1)e+00 &            2.77(4)e+01 &            1.97(2)e+01 \\
              & AK    &  \B 1.90(5)e-01 &  \B -1.06(1)e+00 &  \B 4.43(1)e+00 &  \B 2.53(3)e+01 &  \B 1.77(2)e+01 \\
              & Gibbs &            2.21(8)e-01 &             -7.7(4)e-01 &            4.72(5)e+00 &            2.74(5)e+01 &            1.94(3)e+01 \\
              & DKL   &            2.34(8)e-01 &             -7.1(2)e-01 &            4.78(2)e+00 &            2.82(5)e+01 &            1.98(3)e+01 \\
              \bottomrule
  \end{tabular}%
\end{table*}

\begin{figure*}[tbp]
    \centering
    \subfloat{%
        \resizebox{0.5\linewidth}{!}{%
            \includegraphics[width=\linewidth,trim={5 5 5 5},clip]{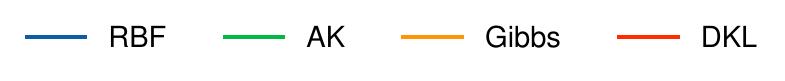}%
        }%
    }\\%
    \addtocounter{subfigure}{-1}%
    \foreach \env in {N17E073,N43W080,N45W123,N47W124}{%
        \subfloat[SMSE in \env\label{fig:\env_random_smse}]{%
            \resizebox{0.25\linewidth}{!}{%
                \includegraphics[width=0.25\linewidth,trim={3 3 3 3},clip]{images/figure_12_\env_SMSE.jpg} %
            }%
        }%
    }\\%
    \foreach \env in {N17E073,N43W080,N45W123,N47W124}{%
        \subfloat[MLSS in \env\label{fig:\env_random_msll}]{%
            \resizebox{0.25\linewidth}{!}{%
                \includegraphics[width=0.25\linewidth,trim={3 3 3 3},clip]{images/figure_12_\env_MSLL.jpg} %
            }%
        }%
    }%
    \caption{\textbf{Random Sampling Metrics versus Number of Collected Samples}.}\label{fig:metrics_random}%
\end{figure*}

\begin{figure*}[tbp]
    \centering
    \setlength{\fboxrule}{5pt}
    \setlength{\fboxsep}{0pt}
    \subfloat[Prediction of RBF in \texttt{N47W124}]{%
        \resizebox{0.25\linewidth}{!}{%
        \fbox{\includegraphics[width=\linewidth]{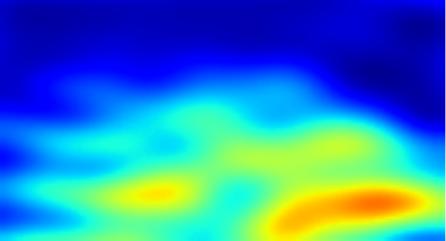}} }
    }
    \subfloat[Prediction of AK in \texttt{N47W124}]{%
        \resizebox{0.25\linewidth}{!}{%
        \fbox{\includegraphics[width=\linewidth]{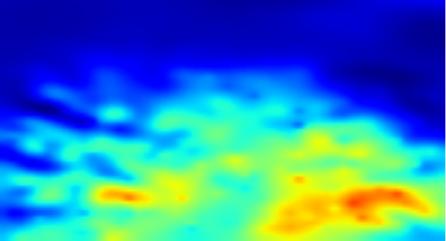}} }
    }
    \subfloat[Prediction of Gibbs in \texttt{N47W124}]{%
        \resizebox{0.25\linewidth}{!}{%
        \fbox{\includegraphics[width=\linewidth]{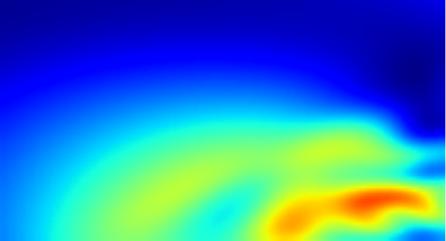}} }
    }
    \subfloat[Prediction of DKL in \texttt{N47W124}\label{fig:viz_random_dkl_mean}]{%
        \resizebox{0.25\linewidth}{!}{%
        \fbox{\includegraphics[width=\linewidth]{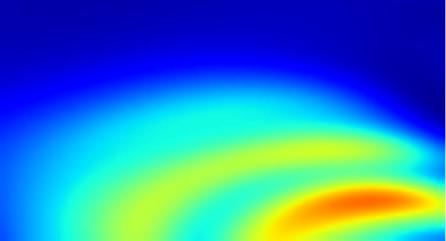}} }
    }\\
    \subfloat[Uncertainty of RBF in \texttt{N47W124}]{%
        \resizebox{0.25\linewidth}{!}{%
        \fbox{\includegraphics[width=\linewidth]{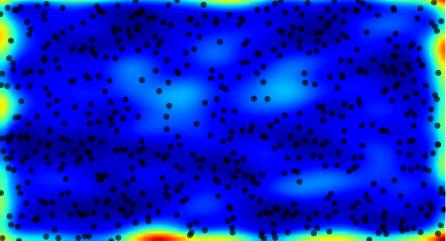}} }
    }
    \subfloat[Uncertainty of AK in \texttt{N47W124}]{%
        \resizebox{0.25\linewidth}{!}{%
        \fbox{\includegraphics[width=\linewidth]{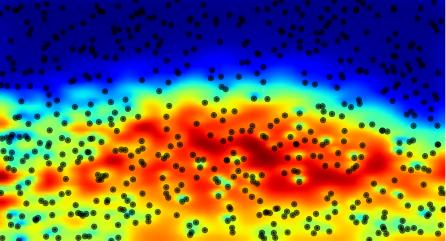}} }
    }
    \subfloat[Uncertainty of Gibbs in \texttt{N47W124}]{%
        \resizebox{0.25\linewidth}{!}{%
        \fbox{\includegraphics[width=\linewidth]{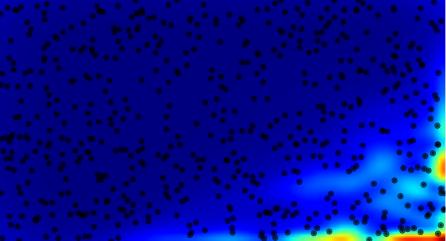}} }
    }
    \subfloat[Uncertainty of DKL in \texttt{N47W124}]{%
        \resizebox{0.25\linewidth}{!}{%
        \fbox{\includegraphics[width=\linewidth]{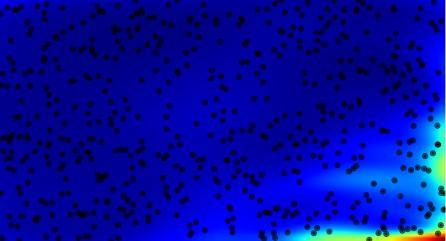}} }
    }\\
    \subfloat[Error of RBF in \texttt{N47W124}]{%
        \resizebox{0.25\linewidth}{!}{%
        \fbox{\includegraphics[width=\linewidth]{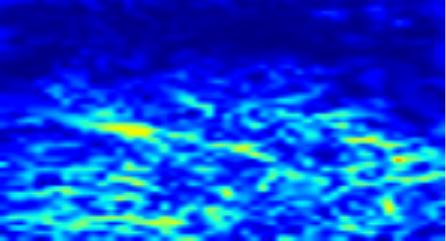}} }
    }
    \subfloat[Error of AK in \texttt{N47W124}]{%
        \resizebox{0.25\linewidth}{!}{%
        \fbox{\includegraphics[width=\linewidth]{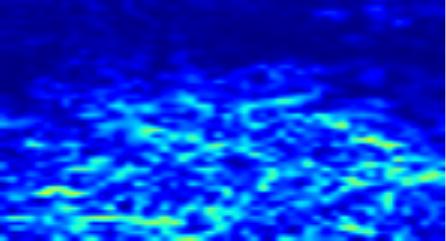}} }
    }
    \subfloat[Error of Gibbs in \texttt{N47W124}]{%
        \resizebox{0.25\linewidth}{!}{%
        \fbox{\includegraphics[width=\linewidth]{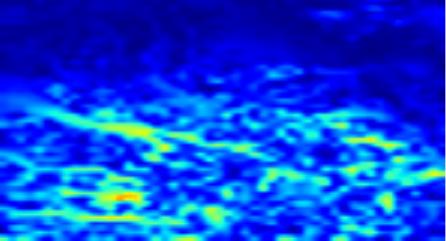}} }
    }
    \subfloat[Error of DKL in \texttt{N47W124}]{%
        \resizebox{0.25\linewidth}{!}{%
        \fbox{\includegraphics[width=\linewidth]{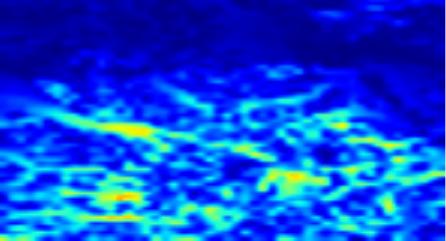}} }
    }\\
    \caption{\textbf{Snapshots of the Random Sampling Experiments with Different Kernels}.}\label{fig:viz_random}%
\end{figure*}

\subsubsection{Evaluation Metrics}\label{sec:metrics}
We care about the prediction performance and whether the predictive uncertainty can effectively reflect the prediction error.
Following standard practice in the GP literature, we use {\em standardized mean squared error (SMSE)} and {\em mean standardized log loss (MSLL)} to measure these quantities~(see Chapter 2.5 in~\cite{rasmussen2005mit}).
SMSE is the mean squared error divided by the variance of test targets.
After this standardization, a trivial method that makes a prediction using the mean of the training targets has an SMSE of approximately $1$.
To take the predictive uncertainty into account, one can evaluate the negative log predictive density~(NLPD), \textit{a.k.a.}, log loss, of a test target,
\begin{equation*}
  -\ln{p(y^{\star}\rvert\mathbf{x}^{\star})}=\frac{\ln(2\pi\nu)}{2}+\frac{(y^{\star}-\mu)^{2}}{(2\nu)},
\end{equation*}
where $\mu$ and $\nu$ are the mean and variance in \Cref{eq:pred_mu,eq:pred_nu}.
MSLL standardizes the log loss by subtracting the log loss obtained under the trivial model, which predicts using a Gaussian with the mean and variance of the training targets.
The MSLL will be approximately zero for naive methods and negative for better methods.
In the experiments, we also measured the root-mean-square error~(RMSE) and the mean absolute error~(MAE).
We report the mean and standard deviation of the metrics over ten runs of the experiments with different random seeds.
For a more obvious quantitative comparison, we present all the benchmarking results in~\Cref{tab:random,tab:active,tab:rig}.
Each number summarizes a metric curve by averaging the curve across the x-axis, \textit{i.e.}, the number of samples, which indicates the averaged \emph{area under the curve}.
A smaller area implies a faster drop in the curve.
For all the metrics, smaller values indicate better performance.

\rowcolors{2}{gray!25}{white}
\begin{table*}[tbp]
  \centering\small
  \caption{\textbf{Active Sampling Benchmarking Results}.}\label{tab:active}
  \begin{tabular}{
      ll
      S[table-format=1.2(3)e-1]
      S[table-format=-1.2(3)e-1]
      S[table-format=1.2(3)]
      S[table-format=1.2(3)e1]
      S[table-format=1.2(3)e1]
    }
    \toprule
    {\textbf{Environment}} & {\textbf{Method}} & {\textbf{SMSE}$\downarrow^{1}_{0}$} & {\textbf{MSLL}$\downarrow^{0}$} & {\textbf{NLPD}$\downarrow$} & {\textbf{RMSE}$\downarrow_{0}$} & {\textbf{MAE}$\downarrow_{0}$} \\
    \midrule
    N17E073   & RBF   &            1.41(4)e-01 &             -9.8(2)e-01 &            4.61(2)e+00 &            2.38(3)e+01 &            1.70(3)e+01 \\
              & AK    &  \B 1.01(2)e-01 &  \B -1.32(4)e+00 &  \B 4.36(2)e+00 &  \B 2.00(2)e+01 &  \B 1.43(2)e+01 \\
              & Gibbs &            1.37(6)e-01 &            -1.20(8)e+00 &            4.59(3)e+00 &            2.35(6)e+01 &            1.72(5)e+01 \\
              & DKL   &            1.33(7)e-01 &            -1.09(5)e+00 &            4.59(3)e+00 &            2.32(6)e+01 &            1.62(5)e+01 \\
    N43W080   & RBF   &            7.8(2)e-02 &            -1.41(1)e+00 &            3.96(1)e+00 &            1.28(1)e+01 &            9.0(1)e+00 \\
              & AK    &  \B 5.1(2)e-02 &  \B -1.72(2)e+00 &  \B 3.74(3)e+00 &  \B 1.02(2)e+01 &  \B 6.9(1)e+00 \\
              & Gibbs &            8.0(6)e-02 &            -1.48(5)e+00 &            3.98(6)e+00 &            1.31(6)e+01 &            9.8(4)e+00 \\
              & DKL   &              7(1)e-02 &             -1.6(1)e+00 &             3.9(1)e+00 &             1.2(1)e+01 &            8.2(6)e+00 \\
    N45W123   & RBF   &            1.47(4)e-01 &             -9.7(1)e-01 &            4.36(1)e+00 &            1.85(2)e+01 &            1.23(2)e+01 \\
              & AK    &  \B 1.08(3)e-01 &  \B -1.55(4)e+00 &  \B 4.16(2)e+00 &  \B 1.57(3)e+01 &  \B 1.14(3)e+01 \\
              & Gibbs &            1.29(6)e-01 &            -1.48(5)e+00 &            4.30(2)e+00 &            1.73(4)e+01 &            1.28(2)e+01 \\
              & DKL   &             1.6(1)e-01 &            -1.18(4)e+00 &            4.35(3)e+00 &            1.91(7)e+01 &            1.35(4)e+01 \\
    N47W124   & RBF   &            2.15(5)e-01 &             -7.5(1)e-01 &            4.75(1)e+00 &            2.70(3)e+01 &            1.90(3)e+01 \\
              & AK    &  \B 1.78(8)e-01 &  \B -1.09(7)e+00 &  \B 4.56(1)e+00 &  \B 2.45(6)e+01 &  \B 1.75(3)e+01 \\
              & Gibbs &            2.04(6)e-01 &             -9.9(5)e-01 &            4.71(2)e+00 &            2.63(4)e+01 &            1.86(3)e+01 \\
              & DKL   &             2.2(1)e-01 &             -8.1(5)e-01 &            4.76(5)e+00 &            2.75(9)e+01 &            1.94(5)e+01 \\
              \bottomrule
  \end{tabular}%
\end{table*}

\begin{figure*}[tbp]
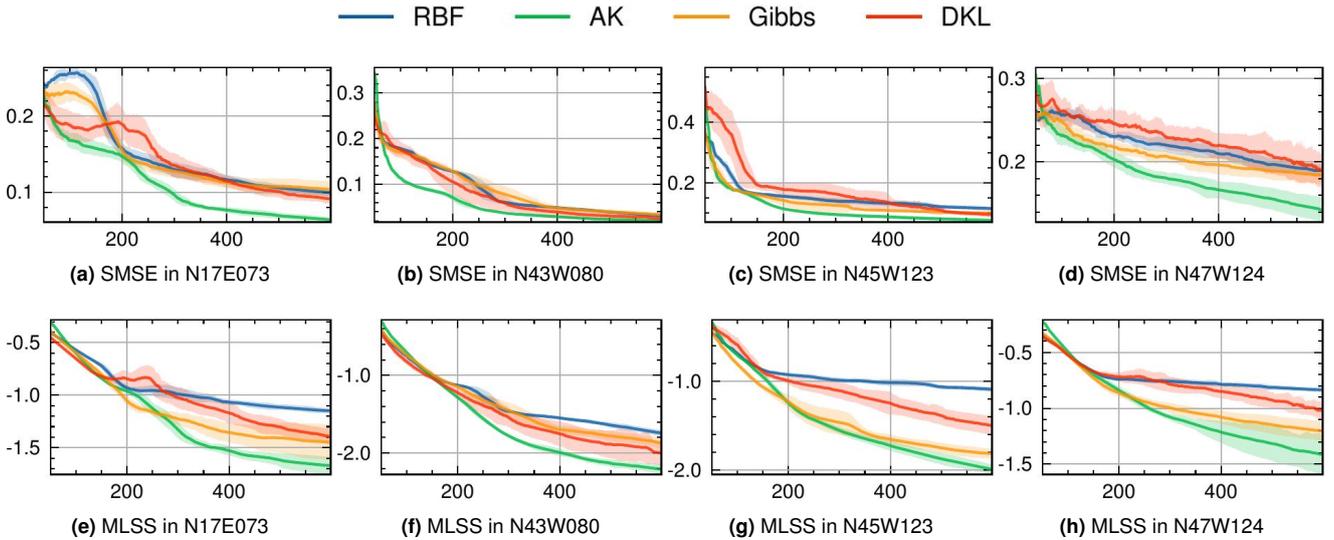

  \centering
  \subfloat{%
    \resizebox{0.5\linewidth}{!}{%
      \includegraphics[width=\linewidth,trim={5 5 5 5},clip]{images/legend_methods.jpg}%
    }%
  }\\%
  \addtocounter{subfigure}{-1}%
  \foreach \env in {N17E073,N43W080,N45W123,N47W124}{%
    \subfloat[SMSE in \env\label{fig:\env_active_smse}]{%
      \resizebox{0.25\linewidth}{!}{%
        \includegraphics[width=0.5\linewidth,trim={3 3 3 3},clip]{images/figure_14_\env_SMSE.jpg} %
      }%
    }%
  }\\%
  \foreach \env in {N17E073,N43W080,N45W123,N47W124}{%
    \subfloat[MLSS in \env\label{fig:\env_active_msll}]{%
      \resizebox{0.25\linewidth}{!}{%
        \includegraphics[width=0.5\linewidth,trim={3 3 3 3},clip]{images/figure_14_\env_MSLL.jpg} %
      }%
    }%
  }%
  \caption{\textbf{Active Sampling Metrics versus Number of Collected Samples}.}\label{fig:metrics_active}%
\end{figure*}

\begin{figure*}[tbp]
  \centering
  \setlength{\fboxrule}{5pt}
  \setlength{\fboxsep}{0pt}
  \subfloat[Prediction of AK in \texttt{N43W080}\label{fig:viz_active_ak_lake_mean}]{%
    \resizebox{0.25\linewidth}{!}{%
    \fbox{\includegraphics[width=\linewidth]{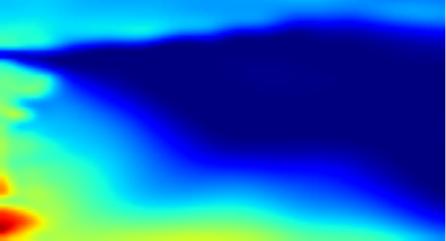}} }
  }
  \subfloat[Prediction of AK in \texttt{N45W123}]{%
    \resizebox{0.25\linewidth}{!}{%
    \fbox{\includegraphics[width=\linewidth]{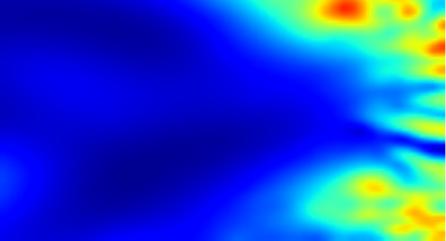}} }
  }
  \subfloat[Prediction of Gibbs in \texttt{N45W123}]{%
    \resizebox{0.25\linewidth}{!}{%
    \fbox{\includegraphics[width=\linewidth]{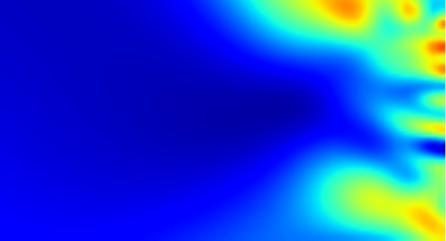}} }
  }
  \subfloat[Prediction of DKL in \texttt{N45W123}\label{fig:viz_active_dkl_mean}]{%
    \resizebox{0.25\linewidth}{!}{%
    \fbox{\includegraphics[width=\linewidth]{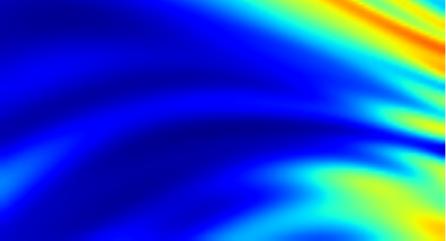}} }
  }\\
  \subfloat[Uncertainty of AK in \texttt{N43W080}\label{fig:viz_active_ak_lake_std}]{%
    \resizebox{0.25\linewidth}{!}{%
    \fbox{\includegraphics[width=\linewidth]{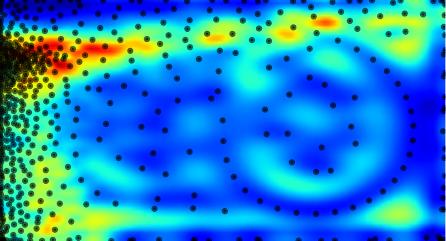}} }
  }
  \subfloat[Uncertainty of AK in \texttt{N45W123}\label{fig:viz_active_ak_river_std}]{%
    \resizebox{0.25\linewidth}{!}{%
    \fbox{\includegraphics[width=\linewidth]{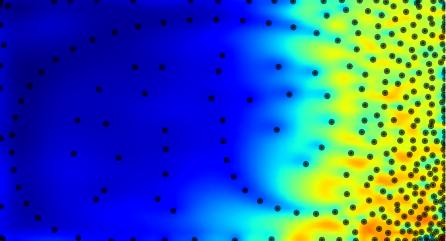}} }
  }
  \subfloat[Uncertainty of Gibbs in \texttt{N45W123}\label{fig:viz_active_gibbs_river_std}]{%
    \resizebox{0.25\linewidth}{!}{%
    \fbox{\includegraphics[width=\linewidth]{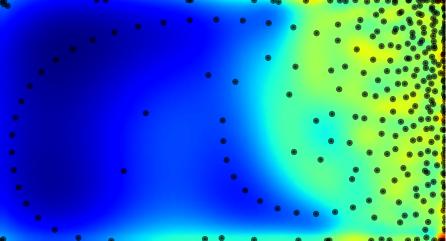}} }
  }
  \subfloat[Uncertainty of DKL in \texttt{N45W123}\label{fig:viz_active_dkl_std}]{%
    \resizebox{0.25\linewidth}{!}{%
    \fbox{\includegraphics[width=\linewidth]{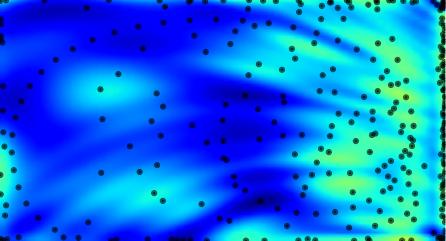}} }
  }\\
  \subfloat[Errorr of AK in \texttt{N43W080}\label{fig:viz_active_ak_lake_error}]{%
    \resizebox{0.25\linewidth}{!}{%
    \fbox{\includegraphics[width=\linewidth]{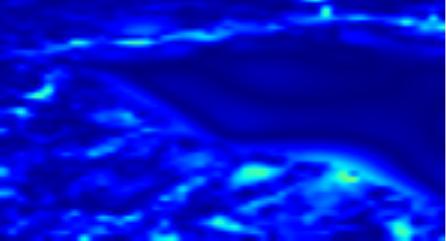}} }
  }
  \subfloat[Error of AK in \texttt{N45W123}\label{fig:viz_active_ak_river_error}]{%
    \resizebox{0.25\linewidth}{!}{%
    \fbox{\includegraphics[width=\linewidth]{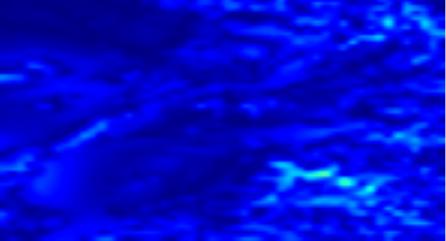}} }
  }
  \subfloat[Error of Gibbs in \texttt{N45W123}\label{fig:viz_active_gibbs_river_error}]{%
    \resizebox{0.25\linewidth}{!}{%
    \fbox{\includegraphics[width=\linewidth]{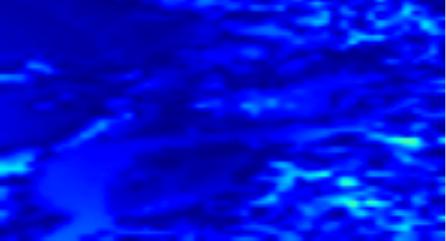}} }
  }
  \subfloat[Error of DKL in \texttt{N45W123}\label{fig:viz_active_dkl_error}]{%
    \resizebox{0.25\linewidth}{!}{%
    \fbox{\includegraphics[width=\linewidth]{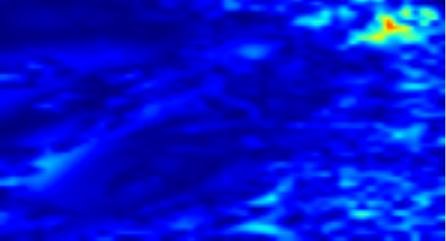}} }
  }\\
  \caption{\textbf{Snapshots of the Active Sampling Experiments with Different Kernels}.}\label{fig:viz_active}%
\end{figure*}

\rowcolors{2}{gray!25}{white}
\begin{table*}[tbp]
    \centering\small
    \caption{\textbf{Robotic Information Gathering Benchmarking Results}.}\label{tab:rig}
    \begin{tabular}{
            ll
            S[table-format=1.2(3)e-1]
            S[table-format=-1.2(3)e-1]
            S[table-format=1.2(3)]
            S[table-format=1.2(3)e1]
            S[table-format=1.2(3)e1]
        }
        \toprule
        {\textbf{Environment}} & {\textbf{Method}} & {\textbf{SMSE}$\downarrow^{1}_{0}$} & {\textbf{MSLL}$\downarrow^{0}$} & {\textbf{NLPD}$\downarrow$} & {\textbf{RMSE}$\downarrow_{0}$} & {\textbf{MAE}$\downarrow_{0}$} \\
        \midrule
        N17E073   & RBF   &            1.45(3)e-01 &             -9.7(2)e-01 &            4.63(2)e+00 &            2.42(2)e+01 &            1.73(2)e+01 \\
                  & AK    &  \B 1.14(4)e-01 &  \B -1.27(3)e+00 &  \B 4.41(4)e+00 &  \B 2.14(4)e+01 &  \B 1.51(2)e+01 \\
                  & Gibbs &            1.43(7)e-01 &            -1.16(4)e+00 &            4.61(4)e+00 &            2.40(7)e+01 &            1.76(6)e+01 \\
                  & DKL   &            1.38(9)e-01 &            -1.01(6)e+00 &            4.61(4)e+00 &            2.38(8)e+01 &            1.67(6)e+01 \\
        N43W080   & RBF   &            7.7(4)e-02 &            -1.40(2)e+00 &            3.94(2)e+00 &            1.27(3)e+01 &             8.8(2)e+00 \\
                  & AK    &  \B 6.6(2)e-02 &  \B -1.64(4)e+00 &  \B 3.78(3)e+00 &  \B 1.14(2)e+01 &  \B 7.69(9)e+00 \\
                  & Gibbs &            7.6(9)e-02 &            -1.50(5)e+00 &            3.91(7)e+00 &            1.25(7)e+01 &             9.0(6)e+00 \\
                  & DKL   &            7.0(1)e-02 &            -1.56(7)e+00 &            3.85(6)e+00 &            1.19(8)e+01 &             8.1(6)e+00 \\
        N45W123   & RBF   &            1.60(6)e-01 &             -9.3(2)e-01 &            4.39(2)e+00 &            1.93(4)e+01 &            1.29(2)e+01 \\
                  & AK    &  \B 1.32(6)e-01 &  \B -1.43(4)e+00 &  \B 4.15(3)e+00 &  \B 1.71(4)e+01 &  \B 1.21(3)e+01 \\
                  & Gibbs &            1.38(7)e-01 &            -1.34(4)e+00 &            4.30(3)e+00 &            1.79(5)e+01 &            1.32(4)e+01 \\
                  & DKL   &             1.7(2)e-01 &            -1.06(8)e+00 &            4.41(6)e+00 &            1.99(9)e+01 &            1.40(6)e+01 \\
        N47W124   & RBF   &            2.23(6)e-01 &             -7.4(1)e-01 &            4.76(1)e+00 &            2.75(3)e+01 &            1.94(2)e+01 \\
                  & AK    &  \B 1.85(4)e-01 &  \B -1.10(3)e+00 &  \B 4.48(3)e+00 &  \B 2.50(3)e+01 &  \B 1.79(3)e+01 \\
                  & Gibbs &            2.12(8)e-01 &             -9.0(5)e-01 &            4.73(3)e+00 &            2.69(5)e+01 &            1.91(2)e+01 \\
                  & DKL   &            2.36(6)e-01 &             -7.7(4)e-01 &            4.78(3)e+00 &            2.83(3)e+01 &            1.99(4)e+01 \\
                  \bottomrule
    \end{tabular}%
\end{table*}

\begin{figure*}[tbp]
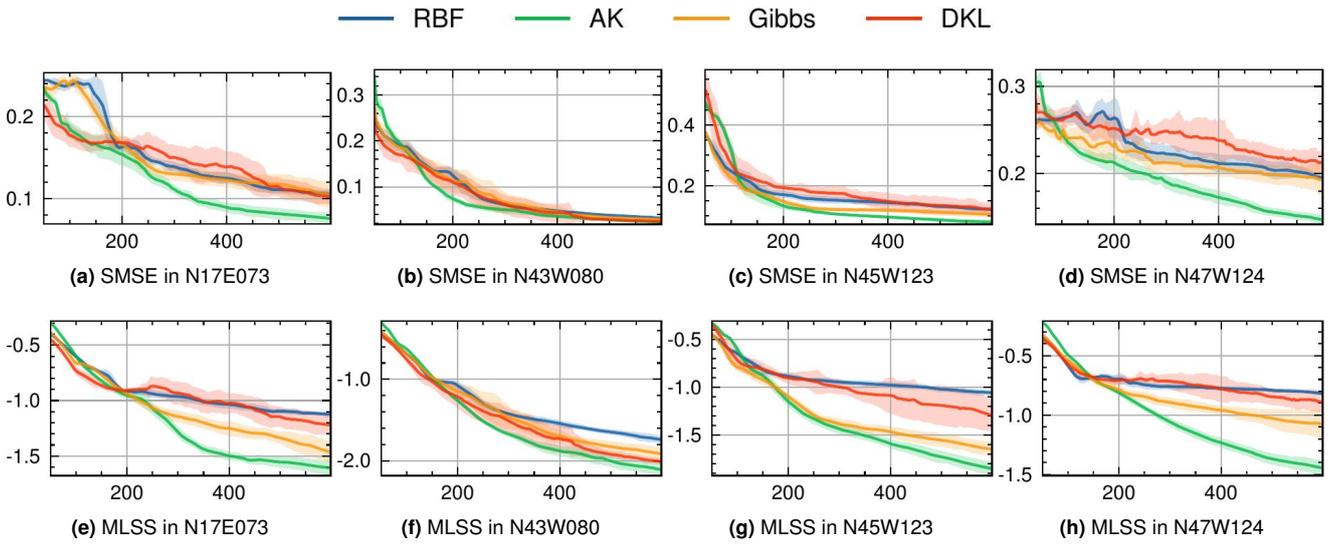

    \centering
    \subfloat{%
        \resizebox{0.5\linewidth}{!}{%
            \includegraphics[width=\linewidth,trim={5 5 5 5},clip]{images/legend_methods.jpg}%
        }%
    }\\%
    \addtocounter{subfigure}{-1}%
    \foreach \env in {N17E073,N43W080,N45W123,N47W124}{%
        \subfloat[SMSE in \env\label{fig:\env_myopic_smse}]{%
            \resizebox{0.25\linewidth}{!}{%
                \includegraphics[width=0.5\linewidth,trim={3 3 3 3},clip]{images/figure_16_\env_SMSE.jpg} %
            }%
        }%
    }\\%
    \foreach \env in {N17E073,N43W080,N45W123,N47W124}{%
        \subfloat[MLSS in \env\label{fig:\env_myopic_msll}]{%
            \resizebox{0.25\linewidth}{!}{%
                \includegraphics[width=0.5\linewidth,trim={3 3 3 3},clip]{images/figure_16_\env_MSLL.jpg} %
            }%
        }%
    }%
    \caption{\textbf{Robotic Information Gathering Metrics versus Number of Collected Samples}.}\label{fig:metrics_myopic}%
\end{figure*}

\begin{figure*}[tbp]
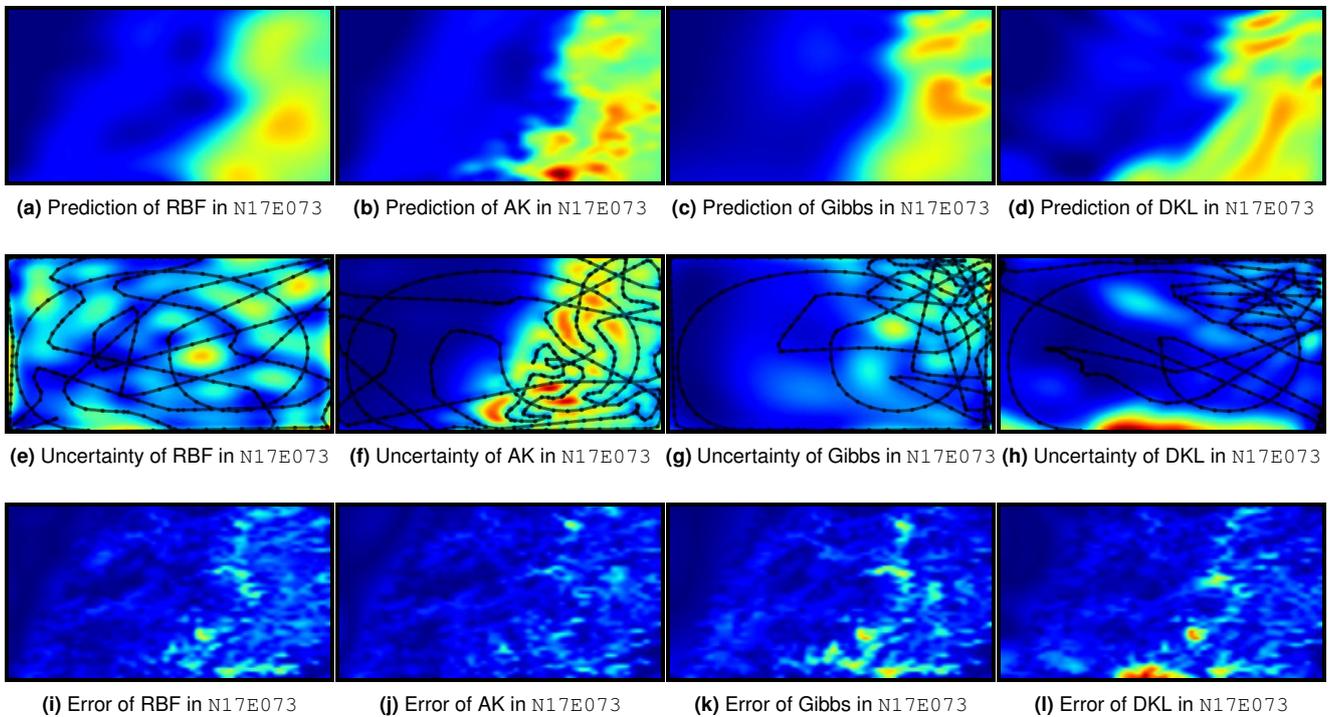

    \centering
    \setlength{\fboxrule}{5pt}
    \setlength{\fboxsep}{0pt}
    \foreach \env in {N17E073}{%
        \foreach \kernel in {RBF, AK, Gibbs, DKL}{%
            \subfloat[Prediction of {\kernel} in \texttt{N17E073}\label{fig:viz_rig_mean_\kernel}]{%
                \resizebox{0.25\linewidth}{!}{%
        \fbox{\includegraphics[width=\linewidth]{images/figure_17_\env_\kernel_mean.jpg}} }}}\\%
        \hspace{0.185\linewidth}
        \foreach \kernel in {RBF, AK, Gibbs, DKL}{%
            \subfloat[Uncertainty of {\kernel} in \texttt{N17E073}\label{fig:viz_rig_std_\kernel}]{%
                \resizebox{0.25\linewidth}{!}{%
        \fbox{\includegraphics[width=\linewidth]{images/figure_17_\env_\kernel_std.jpg}} }}}\\%
        \hspace{0.185\linewidth}
        \foreach \kernel in {RBF, AK, Gibbs, DKL}{%
            \subfloat[Error of {\kernel} in \texttt{N17E073}\label{fig:viz_rig_error_\kernel}]{%
                \resizebox{0.25\linewidth}{!}{%
        \fbox{\includegraphics[width=\linewidth]{images/figure_17_\env_\kernel_error.jpg}} }}}\\%
    }
    \caption{\textbf{Snapshots of the Robotic Information Gathering Experiments with Different Kernels}.}\label{fig:viz_rig}%
\end{figure*}

\begin{figure*}[tbp]
  \centering
  \subfloat{\includegraphics[width=0.8\linewidth,trim={5 5 5 5},clip]{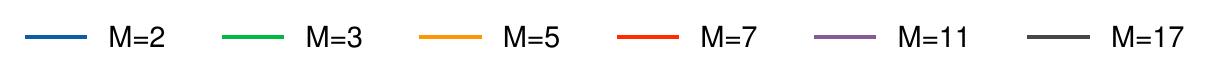}}%
  \addtocounter{subfigure}{-1}\\%
  \foreach \metric in {SMSE,MSLL,MAE}{%
    \foreach \env in {N17E073,N43W080,N45W123,N47W124}{%
      \subfloat[{\metric} in {\env}]{%
        \resizebox{0.25\linewidth}{!}{%
          \includegraphics[width=\linewidth,trim={3 3 3 3},clip]{images/figure_18_\env_\metric.jpg} %
        }%
      }%
    }\\%
  }%
  \caption{\textbf{Sensitivity Analysis of the Number of Base Kernels $M$}.}\label{fig:sensitivity_m}%
\end{figure*}

\begin{figure*}[tbp]
  \centering
  \subfloat{\includegraphics[width=0.8\linewidth,trim={5 5 5 5},clip]{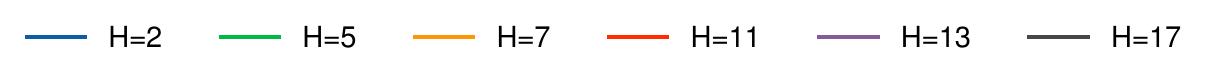}}%
  \addtocounter{subfigure}{-1}\\%
  \foreach \metric in {SMSE,MSLL,MAE}{%
    \foreach \env in {N17E073,N43W080,N45W123,N47W124}{%
      \subfloat[{\metric} in {\env}]{%
        \resizebox{0.25\linewidth}{!}{%
          \includegraphics[width=\linewidth,trim={3 3 3 3},clip]{images/figure_19_\env_\metric.jpg} %
        }%
      }%
    }\\%
  }%
  \caption{\textbf{Sensitivity Analysis of the Number of Hidden Units $H$}.}\label{fig:sensitivity_h}%
\end{figure*}

\begin{figure*}[tbp]
  \centering
  \subfloat{\includegraphics[width=\linewidth,trim={5 5 5 5},clip]{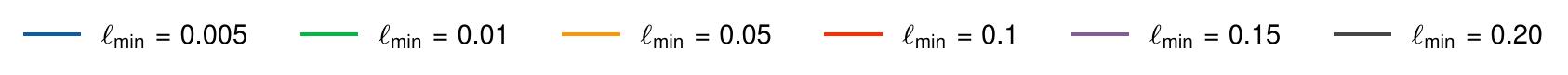}}%
  \addtocounter{subfigure}{-1}\\%
  \foreach \metric in {SMSE,MSLL,MAE}{%
    \foreach \env in {N17E073,N43W080,N45W123,N47W124}{%
      \subfloat[{\metric} in {\env}]{%
        \resizebox{0.25\linewidth}{!}{%
          \includegraphics[width=\linewidth,trim={3 3 3 3},clip]{images/figure_20_\env_\metric.jpg} %
        }%
      }%
    }\\%
  }%
  \caption{\textbf{Sensitivity Analysis of the Minimum Primitive Length-Scale $\ell_{\text{min}}$}.}\label{fig:sensitivity_min}%
\end{figure*}

\begin{figure*}[tbp]
  \centering
  \subfloat{\includegraphics[width=\linewidth,trim={5 5 5 5},clip]{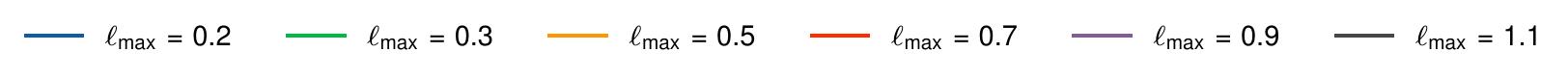}}%
  \addtocounter{subfigure}{-1}\\%
  \foreach \metric in {SMSE,MSLL,MAE}{%
    \foreach \env in {N17E073,N43W080,N45W123,N47W124}{%
      \subfloat[{\metric} in {\env}]{%
        \resizebox{0.25\linewidth}{!}{%
          \includegraphics[width=\linewidth,trim={3 3 3 3},clip]{images/figure_21_\env_\metric.jpg} %
        }%
      }%
    }\\%
  }%
  \caption{\textbf{Sensitivity Analysis of the Maximum Primitive Length-Scale $\ell_{\text{max}}$}.}\label{fig:sensitivity_max}%
\end{figure*}

\begin{figure*}[tbp]
  \centering
  \subfloat{\includegraphics[width=0.5\linewidth,trim={5 5 5 5},clip]{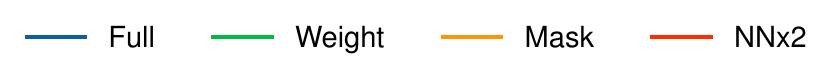}}%
  \addtocounter{subfigure}{-1}\\%
  \foreach \metric in {SMSE,MSLL,MAE}{%
    \foreach \env in {N17E073,N43W080,N45W123,N47W124}{%
      \subfloat[{\metric} in {\env}\label{fig:ablation_\metric_\env}]{%
        \resizebox{0.25\linewidth}{!}{%
          \includegraphics[width=\linewidth,trim={3 3 3 3},clip]{images/figure_22_\env_\metric.jpg} %
        }%
      }%
    }\\%
  }%
  \caption{\textbf{Results of the Four Variants in the Ablation Study}.}\label{fig:ablation}%
\end{figure*}

\begin{figure*}[tbp]%
  \centering
  \subfloat{\includegraphics[width=0.5\linewidth,trim={5 5 5 5},clip]{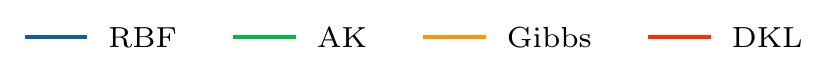}}%
  \addtocounter{subfigure}{-1}\\%
  \subfloat[Training MSLL in Volcano\label{fig:volcano_train_msll}]{%
    \resizebox{0.25\linewidth}{!}{%
  \includegraphics[width=\linewidth,height=0.6\linewidth]{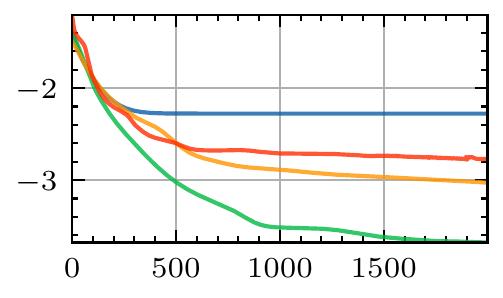} }}%
  \subfloat[Test MSLL in Volcano\label{fig:volcano_test_msll}]{%
    \resizebox{0.25\linewidth}{!}{%
  \includegraphics[width=\linewidth,height=0.6\linewidth]{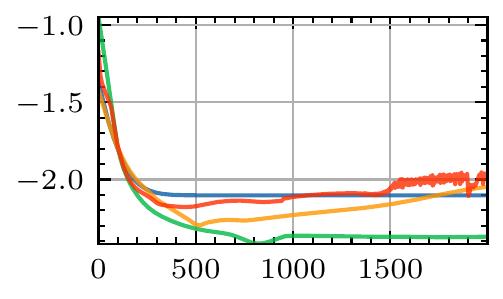} }}%
  \subfloat[Training MSLL in N17E073\label{fig:N17E073_train_msll}]{%
    \resizebox{0.25\linewidth}{!}{%
  \includegraphics[width=\linewidth,height=0.6\linewidth]{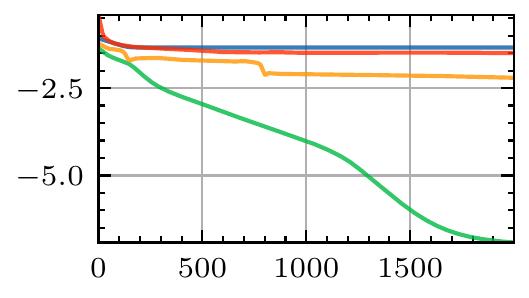} }}%
  \subfloat[Test MSLL in N17E073\label{fig:N17E073_test_msll}]{%
    \resizebox{0.25\linewidth}{!}{%
  \includegraphics[width=\linewidth,height=0.6\linewidth]{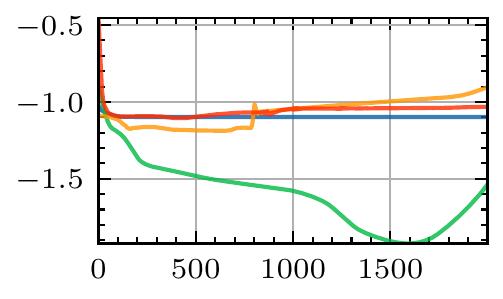}}}\\%
  \caption{\textbf{Results of the Over-Fitting Analysis in the \textit{Volcano} Environment Introduced in \Cref{fig:volcano_env} and \texttt{N17E073}}.}\label{fig:overfitting}%
\end{figure*}

\begin{figure*}[tbp]
  \centering
  \subfloat[An Autonomous Surface Vehicle is performing elevation information gathering task using a sonar.\label{fig:field_demo}]{\includegraphics[width=\linewidth,height=0.55\linewidth]{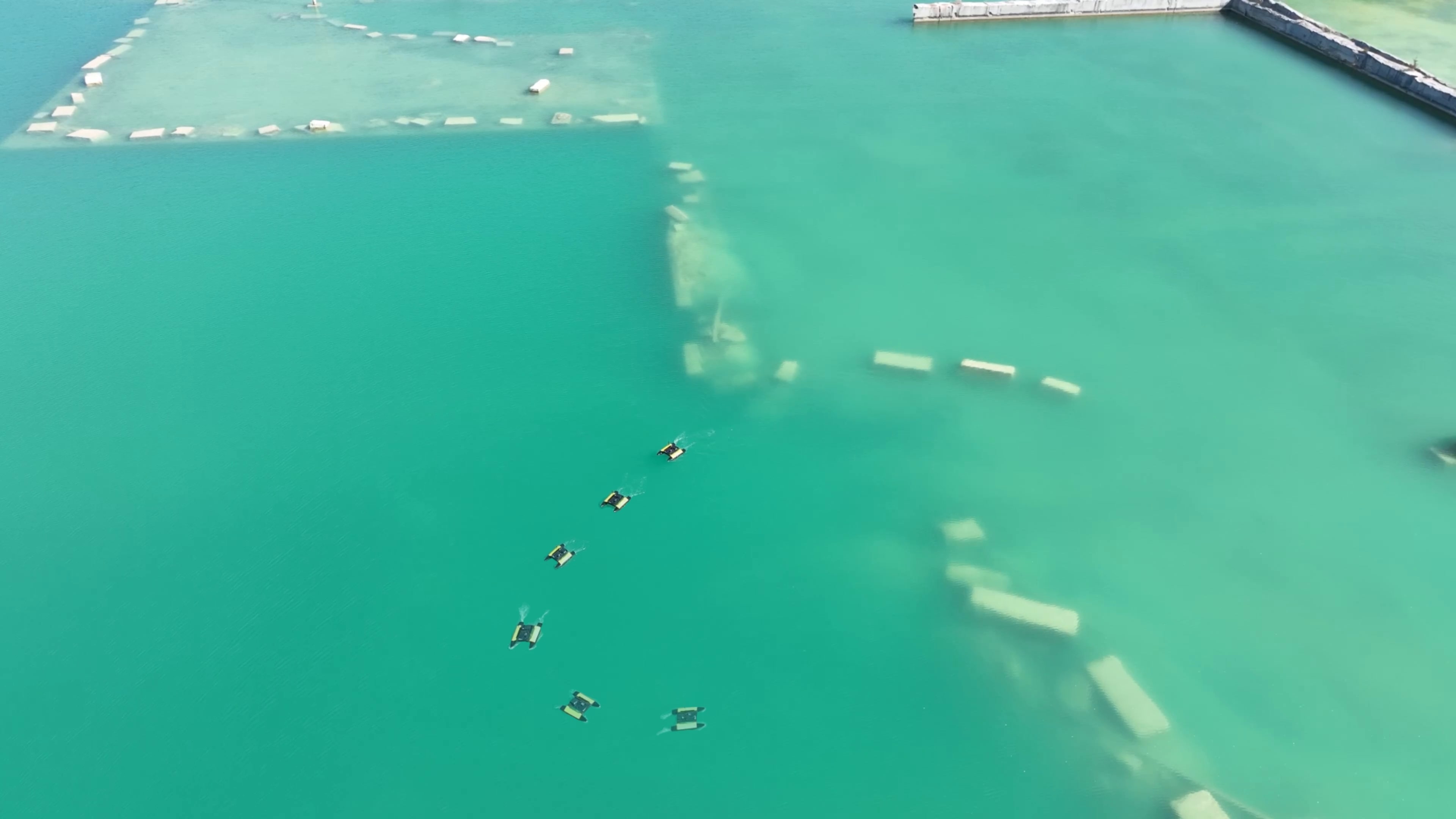}}\\%
  \subfloat[System\label{fig:field_system}]{%
    \resizebox{0.5\linewidth}{!}{%
      \begin{tikzpicture}
        \node(a){\includegraphics[width=\linewidth,height=0.55\linewidth]{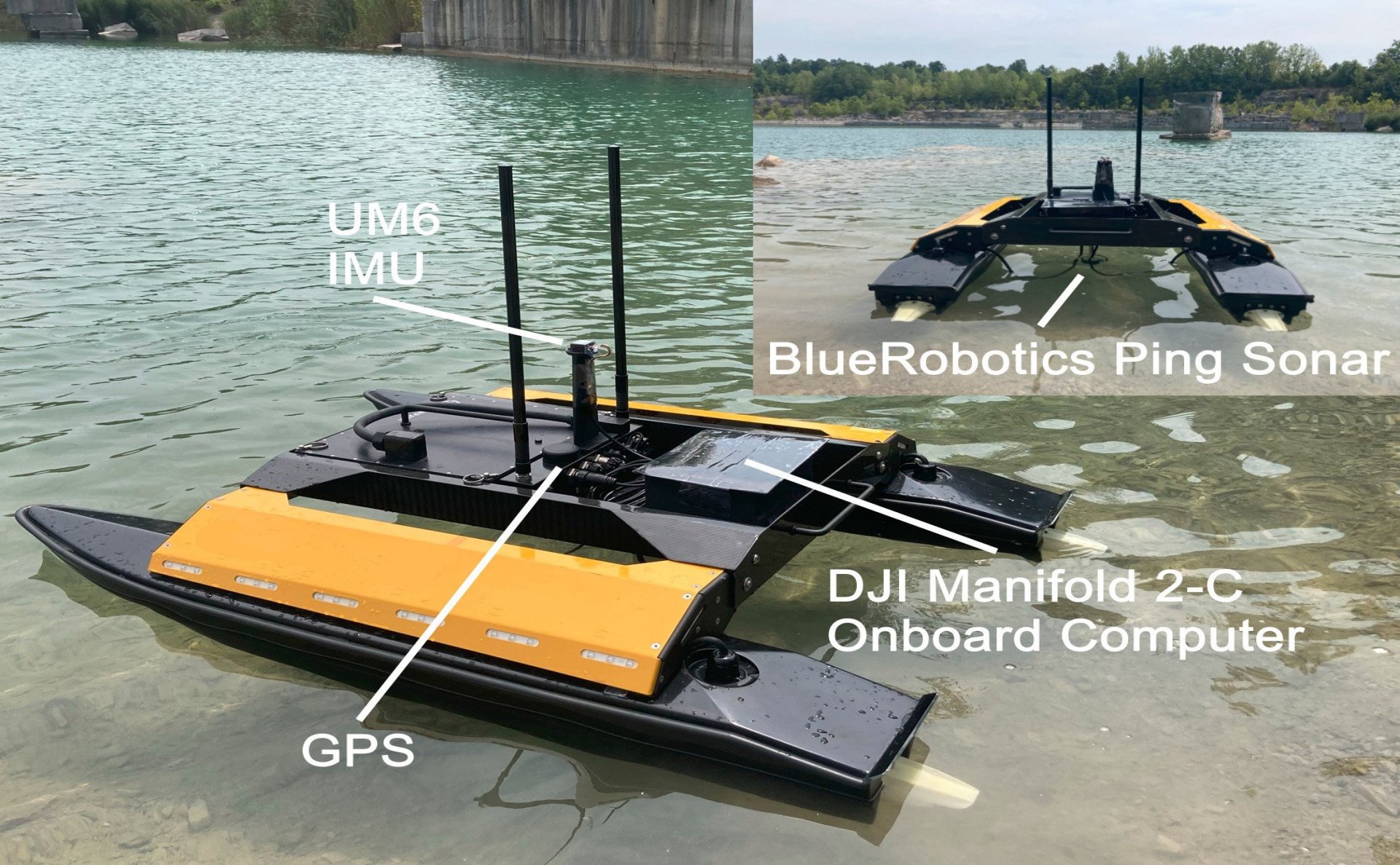} };
      \end{tikzpicture}%
  }}%
  \subfloat[Workspace\label{fig:field_workspace}]{%
    \resizebox{0.5\linewidth}{!}{%
      \begin{tikzpicture}%
        \node(a){\includegraphics[width=\linewidth,height=0.55\linewidth]{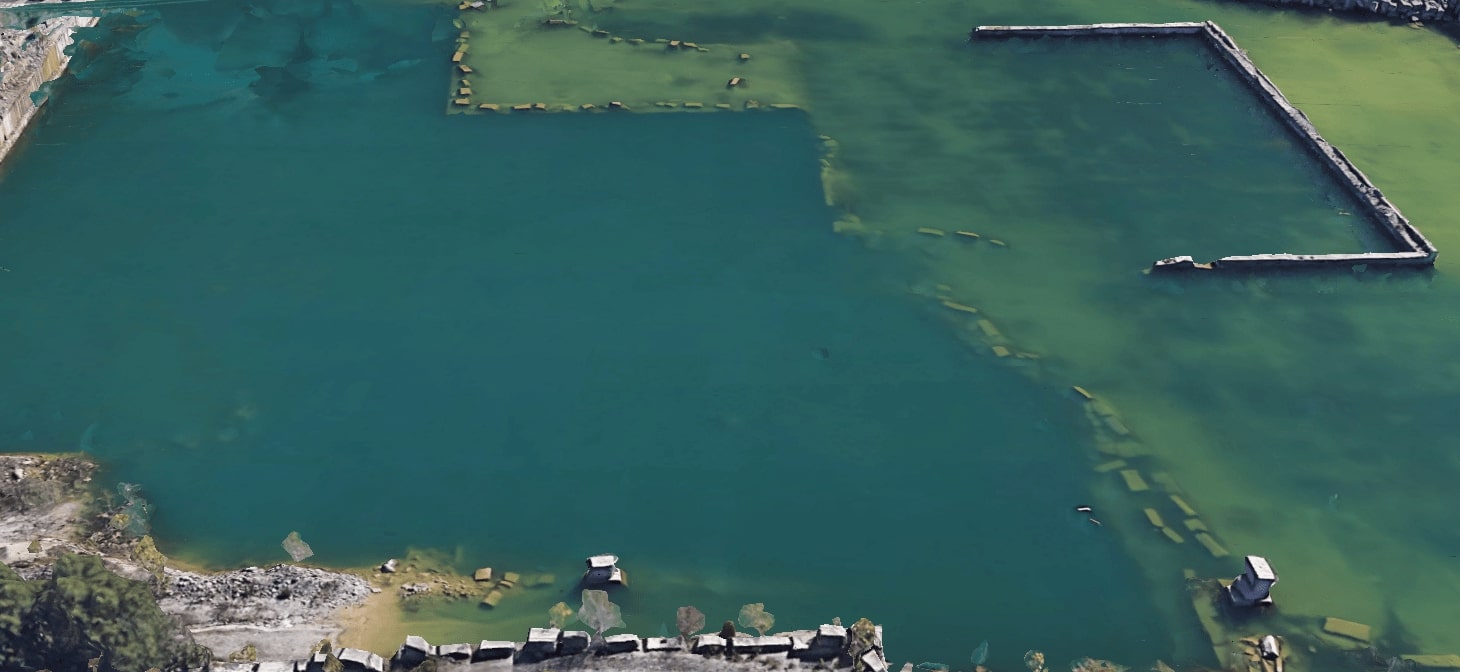} };%
        \node at(a.center)[draw,orange,line width=2,ellipse,minimum width=40,minimum height=40,rotate=0,yshift=-40pt,xshift=-160pt]{};%
        \node at(a.center)[draw,orange,line width=2,rectangle,minimum width=30,minimum height=50,rotate=0,yshift=50pt,xshift=35pt]{};%
        \node at(a.center)[draw,white,line width=2,rectangle,minimum width=240,minimum height=160,rotate=0,yshift=0pt,xshift=-65pt]{};%
        \draw [red,-stealth,line width=3](-4,-3.5) -- (-4,-2.5);%
        \draw [green,-stealth,line width=3](-4,-3.5) -- (-5,-3.5);%
        \node [white] at (-4,-2) {\Huge $x$};%
        \node [white] at (-5.5,-3.3) {\Huge $y$};%
  \end{tikzpicture}}}\\
  \subfloat[Early Snapshot\label{fig:field_snapshot_early}]{%
    \resizebox{0.5\linewidth}{!}{%
      \begin{tikzpicture}
        \node(a){\includegraphics[width=\linewidth,height=0.6\linewidth]{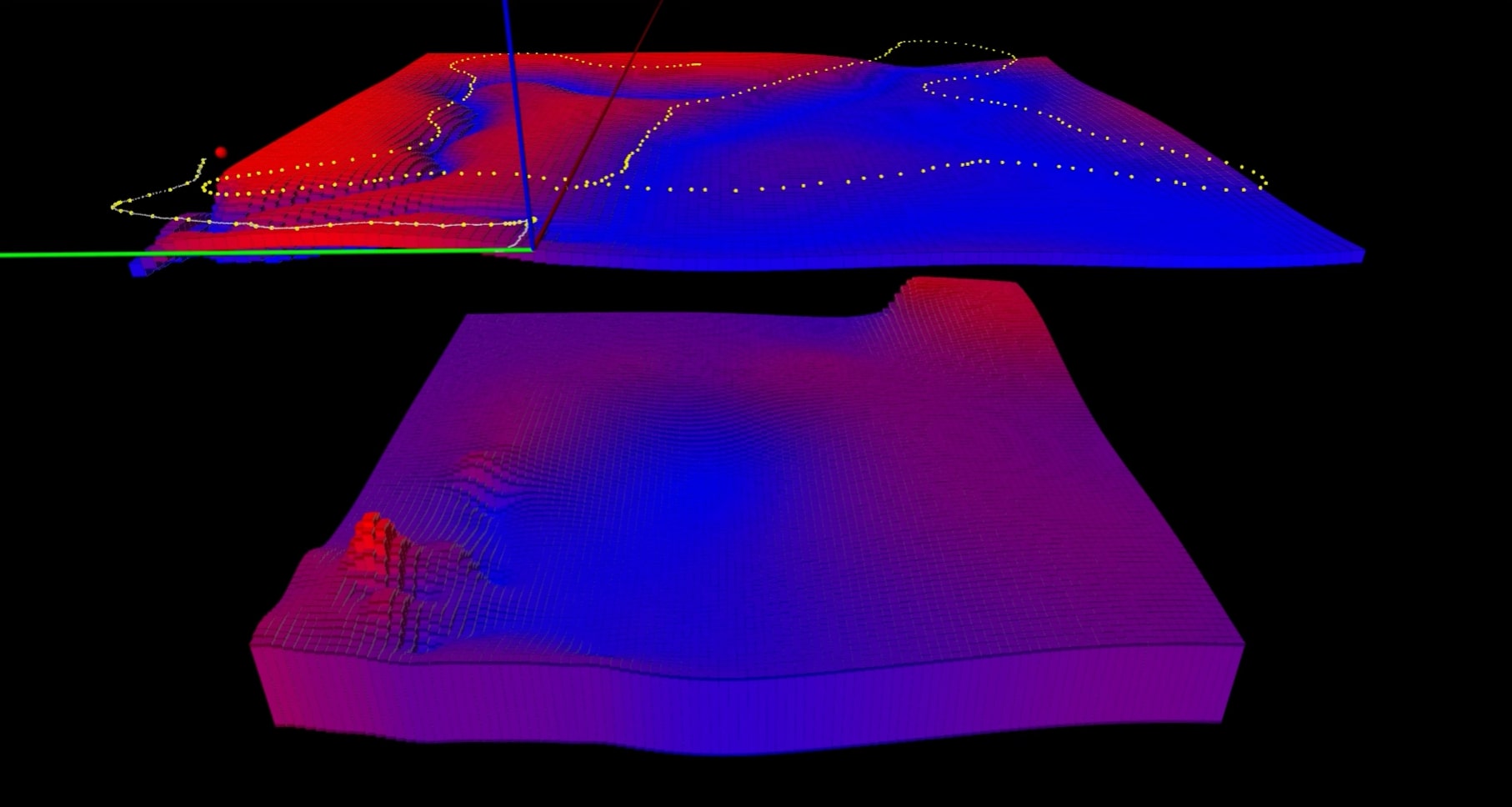} };%
      \end{tikzpicture}
  }}%
  \subfloat[Final Snapshot\label{fig:field_snapshot_final}]{%
    \resizebox{0.5\linewidth}{!}{%
      \begin{tikzpicture}
        \node(a){\includegraphics[width=\linewidth,height=0.6\linewidth]{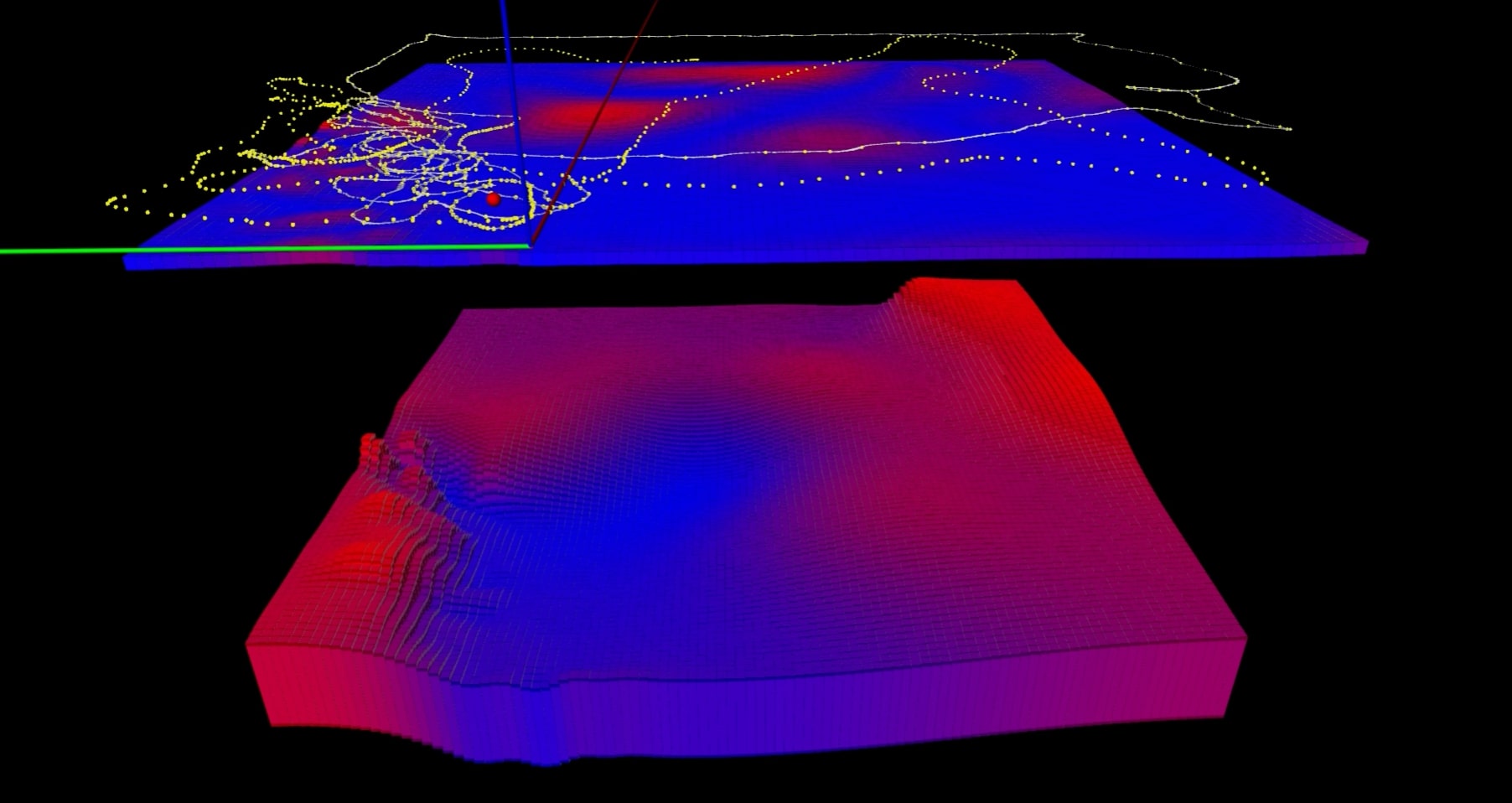}};%
        \node at(a.center)[draw,orange,line width=2,ellipse,minimum width=100,minimum height=100,rotate=0,yshift=-45pt,xshift=-110pt]{};%
        \node at(a.center)[draw,orange,line width=2,rectangle,minimum width=80,minimum height=80,rotate=0,yshift=10pt,xshift=80pt]{};%
        \draw [white,-stealth,line width=2](7,3.5) -- (5.5,4.2);%
        \node [white] at (6.5,3.2) {\LARGE Collected Data};%
        \draw [white,-stealth,line width=2](-2.5,2) -- (-2.9,2.5);%
        \node [white] at (-2.5,1.7) {\LARGE Informative Waypoint};%
        \draw [white,-stealth,line width=2](-2,4.3) -- (-2,3.9);%
        \node [white] at (-2,4.5) {\LARGE High Uncertainty};%
        \draw [white,-stealth,line width=2](4,2.5) -- (3.3,2.8);%
        \node [white] at (4.3,2.1) {\LARGE Low Uncertainty};%
        \draw [white,-stealth,line width=2](-5.2,0.2) -- (-4.8,-0.4);%
        \node [white] at (-5.2,0.5) {\LARGE High Elevation};%
        \draw [white,-stealth,line width=2](-0.5,-2) -- (-1,-1.5);%
        \node [white] at (-0.2,-2.5) {\LARGE Low Elevation};%
  \end{tikzpicture}}}%
  \caption{\textbf{An Active Elevation Mapping Field Experiment}.
    \textbf{(a)} illustrates the physical space the ASV is mapping, and \textbf{(b)} shows the ASV and its components. \textbf{(c)} shows the rectangular workspace for the elevation mapping experiment. We can see two areas with significant elevation features in two highlighted areas, but other regions are opaque. \textbf{(d)} and \textbf{(f)} are two snapshots of the GPR prediction in the rectangular workspace, with the predictive-mean map at the bottom and the uncertainty map (\textit{i.e.}, standard deviation) at the top. In \textbf{(d)}, the lower part shows that some features of the highlighted areas have already been detected. The uncertainty is significant on the left side of the workspace and a smaller region in the top right. \textbf{(e)} shows the snapshot at the end of the experiment. The ASV has extensively explored the lower left portion and has a detailed estimate of its elevation map. The smooth portion in the middle shows differences in elevation, which are not visible in the satellite image. The remaining areas of high uncertainty are at boundaries of elevation changes in that region and the top right.
  }\label{fig:field_snapshot}
\end{figure*}

\subsubsection{Random Sampling Results}\label{sec:random_sampling}
\Cref{tab:random} gives a positive answer to \textbf{Q1} firmly.
The AK consistently outperforms other kernels across all the considered environments and evaluation metrics.
To avoid clutter, we only visualize the SMSE and MLSS curves because they are normalized versions of RMSE and NLPD and the results of MAE are consistent with that of RMSE.

\Cref{fig:metrics_random} shows the metrics versus the number of collected samples of the four kernels in all environments.
From the SMSE curves, we can see that the advantage of the AK (\textit{i.e.}, the green line) is most significant in \texttt{N47W124}, followed by \texttt{N17E073} and then \texttt{N45W123}.
This order complies with the changes in the spatial variability of these environments.
In environment \texttt{N43W080}, all the lines are overlapped.
\texttt{N43W080} is the environment that has two spots with drastic variations.
Too few random samples landed on the two spots to allow the AK to learn better prediction.
That said, the MLSS curve of the AK is still outstanding in this environment.
The advantage of the AK on uncertainty quantification is significant in all environments.
The Gibbs kernel also has better uncertainty quantification than the RBF kernel and DKL.

\Cref{fig:viz_random} visually compares kernels' prediction, uncertainty, and absolute error after collecting $570$ samples in environment \texttt{N47W124}.
Note that the prediction and error maps use the same color scale for easy comparison across different methods.
Each uncertainty map uses its color scale -- red color only indicates relatively high uncertainty \emph{within} the map.
These rules applied to other heat maps hereafter.
The AK learns more detailed environmental features~(\textit{c.f.},~\Cref{fig:env_N47W124}), hence obtaining better SMSE; the AK also assigns higher uncertainty to the region that is relatively more difficult to model, thus giving better MSLL.
As a comparison, the RBF kernel ignores these details and assigns higher uncertainty to the sparsely sampled areas.
The Gibbs kernel also has a smooth prediction in the complex region because it learns an incorrect length-scale function.
Instead of assigning small length-scales to the complex region, it places them in the lower-right corner, indicated by the high uncertainty.
DKL's prediction and uncertainty maps have similar patterns to the Gibbs kernel.

\subsubsection{Active Sampling Results}\label{sec:active_sampling}
The objective of active sampling experiments is to investigate whether prediction uncertainty can influence sampling towards significant areas and ultimately enhance accuracy.
By comparing the SMSE results of the AK in~\Cref{tab:random} and~\Cref{tab:active}, we observe a clear improvement in accuracy when using the active sampling strategy.
Specifically, the relative accuracy improvements are $9\%, 15\%, 23\%$, and $6\%$ in \texttt{N17E073, N43W080, N45W123}, and \texttt{N47W124}, respectively, which answers \textbf{Q2}.
The AK's better uncertainty quantification can further enhance prediction accuracy when the data collection strategy is guided by predictive uncertainty.
However, we do not observe consistent improvements when using active sampling with the other kernels.
Although they all improve the SMSE in \texttt{N45W123} and \texttt{N47W124}, they do not improve the accuracy in the other two environments.
Note that the relative improvements in \texttt{N17E073} and \texttt{N47W124} are smaller because the AK has already achieved good accuracy in these two environments when using random samples, so there is less room to improve than in the other two environments.


The AK still performs the best in the active sampling experiments, as seen in \Cref{tab:active} and \Cref{fig:metrics_active}.
The SMSE curves in \Cref{fig:metrics_active} and \Cref{fig:metrics_random} are similar, except that the advantage gap of the AK shrinks in \texttt{N47W124} and increases in \texttt{N43W080}.
We attribute the faster error drop in \texttt{N43W080} to the better sample distribution.
\Cref{fig:viz_active_ak_lake_mean,fig:viz_active_ak_lake_std} show that, when using the AK, more informative samples are collected in the complex regions in \texttt{N43W080}.
\Cref{fig:viz_active} also shows the prediction, uncertainty, and $570$ samples of the three non-stationary kernels in \texttt{N45W123}, where all methods provide better accuracy when using active sampling strategies.
The predictions of the AK and the Gibbs kernel are visually similar.
The minor difference is located at the lower-right corner, where the AK learns more details~(\textit{c.f.},~\Cref{fig:env_N45W123}).
This difference comes from the different sampling patterns.
The AK samples the right part densely while the Gibbs kernel emphasizes the upper-right~(\textit{c.f.},~\Cref{fig:viz_active_ak_river_std,fig:viz_active_gibbs_river_std}).
Also, the Gibbs kernel samples the left part of the environment very sparsely.
\Cref{fig:viz_active_dkl_mean} shows that DKL is good at depicting the river.
However, it connects the two ``hotspots'' at the upper-right corner, which is an interesting phenomenon: two non-adjacent locations are correlated.
This phenomenon can be found in all the DKL predictions~(see \Cref{fig:viz_random_dkl_mean,fig:viz_rig_mean_DKL}).
The cause of this behavior is that the neural network in DKL warps the geometry of the input space, so the correlation of two given data points is no longer proportional to their distance in the original input space.
It is non-trivial to explain the prediction uncertainty and sampling distribution of DKL shown in~\Cref{fig:viz_active_dkl_std}.

\subsubsection{Informative Planning Results}\label{sec:informative_planning}
The RIG experiments are more challenging than random and active sampling because once the robot decides to visit an informative waypoint, it has to collect the intermediate samples along the trajectory, so the results in~\Cref{tab:rig} should not be compared with that of~\Cref{tab:random,tab:active}.
Given a fixed maximum number of samples, the number of decision epochs of RIG is much smaller than that of active sampling, which makes informed decisions more essential.
\Cref{tab:rig} shows that AK consistently leads across all metrics in the four environments with the simple informative planning strategy described in \Cref{alg:rig_strategy}.
The conclusions we can draw from~\Cref{fig:metrics_myopic} are the same as in active sampling experiments.
From~\Cref{fig:metrics_myopic}, we can see again that the AK has the fastest error reduction, especially in \texttt{N47W124}.
All non-stationary kernels have better MLSS than the stationary baseline.
The AK ranks first in MSLL, and the Gibbs kernel outperforms DKL.

\Cref{fig:viz_rig} is a snapshot of different methods' prediction, uncertainty, and absolute error after collecting $400$ samples in \texttt{N17E073}.
The prediction maps show that the RBF kernel misses many environmental features that non-stationary kernels can capture.
We observe the following behaviors by comparing the patterns in the uncertainty maps and error maps.
\begin{itemize}
  \item Regardless of the prediction errors, the RBF kernel gives the less-sampled area higher uncertainty, so the robot's sampling path uniformly covers the space.
  \item The AK assigns higher uncertainty in the regions with more significant spatial variation; thus, the sampling path focuses more on the complex region.
  \item The Gibbs kernel also has higher uncertainty in the rocky region but does not assign high uncertainty to the lower right.
    Therefore, the sampling path concentrates on the upper-right corner and misses some high-error spots at the bottom.
  \item When using DKL, the robot also samples the upper-right corner densely, and the prediction error at the bottom of the map is the largest across different methods.
  However, DKL places high uncertainty in the high-area region, which can guide the robot to visit these spots later.
\end{itemize}

\subsection{Further Evaluation and Analysis}\label{sec:further_analysis}
We evaluate the AK under different parameter settings for sensitivity analysis and compare four variants of the AK for ablation study.
The challenges of learning the model are also discussed in this section.

\subsubsection{Sensitivity Analysis}\label{sec:sensitivity}
We use the same experiment configurations as the main experiments in the sensitivity analysis but only run the random sampling strategy.
In each analysis, we only change one target parameter to different settings and keep all the other parameters fixed.
\Cref{fig:sensitivity_m} presents the sensitivity analysis results of the number of base kernels $M$, which should be larger than $2$.
Increasing $M$ brings better performance, albeit with a diminishing return and higher computational complexity.
Choosing a number in the range of $[5, 10]$ is a good trade-off between performance and computational efficiency.

\Cref{fig:sensitivity_h} shows that the AK is not sensitive to the number of hidden units in the neural network as long as $H$ is not too small.
When $H=2$, the uncertainty quantification ability decreases, as indicated by the blue MSLL curve.
In this case, the AK can only blend the minimum and maximum primitive length-scales, and the instance selection mechanism can only use a two-dimensional membership vector.

Smaller $\ell_{\text{min}}$ yields better performance, as shown in~\Cref{fig:sensitivity_min}, albeit with a diminishing improvement.
The blue and green lines overlap, meaning that the advantage is negligible when choosing a minimum length-scale smaller than $0.01$.
If the inputs are normalized to $[-1, 1]$, setting the minimum primitive length-scale to $0.01$ is appropriate.
It is worth noting that this is the minimum primitive length-scale for the length-scale selection component.
It does not mean that the AK can only learn the minimum correlation corresponding to this minimum length-scale because the instance selection component can further decrease the kernel values.

As shown in \Cref{fig:sensitivity_max}, the AK is robust to the choice of the maximum length-scale as long as it is not too small, \textit{e.g.}, $0.2$ or $0.3$.
If the inputs are normalized to $[-1, 1]$, choosing a value in the range $[0.5, 1.0]$ is reasonable.

To conclude, these results are positive indicators of addressing \textbf{Q3}: the AK has robust performance to various parameter settings and does not require laborious parameter tuning.

\subsubsection{Ablation Study}\label{sec:ablation}
We compare four AK variants in the ablation study via random sampling experiments.
$\mathtt{Full}$ means the AK presented in the paper, $\mathtt{Weight}$ represents the AK with only length-scale selection, $\mathtt{Mask}$ stands for instance selection alone, and $\mathtt{NNx2}$ uses two separated neural networks to parameterize the similarity attention and visibility attention independently.
\Cref{fig:ablation} shows that using only the instance selection component deteriorates the performance significantly, so the length-scale selection component contributes more to the performance, which answers \textbf{Q4}.
We do not observe obvious performance change after dropping the instance selection component.
Nonetheless, as illustrated in \Cref{fig:1d_ak}, we expect instance selection to provide better modeling of sharp transitions.
Since instance selection improves the prediction only in a small region, the improvement might be subtle in the aggregated evaluation metrics.
With our current training scheme, using two separate neural networks does not provide a better performance, and one of the MSLL curves is surpassed by the one-network version~(\Cref{fig:ablation_MSLL_N43W080}).
The two-network implementation might show its strength with a more refined approach to parameter training.

\subsubsection{Over-Fitting Analysis}\label{sec:overfitting}
Non-stationary kernels can enhance the modeling flexibility of GPR, but they are also more susceptible to over-fitting. This can lead to degraded prediction accuracy and uncertainty estimates.
To evaluate the robustness of non-stationary kernels,
we present an over-fitting analysis in \texttt{N17E073} and the Mount St. Helens environment.
The latter is referred to as the \texttt{volcano} environment hereafter.
We sample $600$ training data from the environment uniformly at random.
All the training configurations are the same as in~\Cref{sec:models}, except for the number of optimization iterations.
We train all the models for $2000$ iterations and evaluate the prediction on the training set and a $100\times{100}$ test grid at each optimization step.
\Cref{fig:overfitting} shows the training and test MSLL.
In some environments, as shown in~\Cref{fig:volcano_train_msll,fig:volcano_test_msll}, the AK is fairly robust, while the Gibbs kernel and DKL show a clear over-fitting trend -- the training MSLL goes down while the test MSLL goes up.
However, as shown in \Cref{fig:N17E073_train_msll,fig:N17E073_test_msll}, all the non-stationary kernels suffer from over-fitting in some environments, such as \texttt{N17E073.}
To mitigate this issue, after collecting one new sample, the optimizer takes only one gradient step on the whole dataset.
This heuristic training scheme works well in practice.
We have tried to optimize the model for more iterations at each decision epoch.
All the non-stationary kernels give poor prediction (the AK is still more robust in this case), and the issue persists even after collecting more data.
Overall, the answer to \textbf{Q5} is positive: the AK is more robust to over-fitting than other non-stationary kernels, but it can still over-fit in some environments.
Developing more advanced training schemes to mitigate over-fitting is an essential future direction.

\subsection{Field Experiment}\label{sec:field}
The proposed AK is demonstrated in a RIG task -- active elevation, \textit{a.k.a.} bathymetric mapping for underwater terrain.
\Cref{fig:field_demo} shows our robot working in the environment.
The goal is to explore an \textit{a priori} unknown quarry lake and build an elevation map of the underwater terrain.
There are two reasons for choosing this task.
First, the underwater terrain is static, so the ground-truth environment is available by aggregating the sampled data across different field experiment trials after offsetting the water surface level.
Second, the underwater terrain in our target environment has a clear separation between ``interesting'' regions and ``boring'' areas, which makes it an ideal testbed for RIG with non-stationary GPs.

\subsubsection{Target Environment}
The target environment is a quarry lake formed by seeped-in groundwater and precipitation since mining and quarrying have been suspended for a long time.
The floor of the quarry lake is complex in that there are many submerged quarry stones and even abandoned equipment.
Our goal is to build an elevation map within the workspace, \textit{i.e.}, the white rectangle shown in \Cref{fig:field_workspace}, with a small number of samples.
The workspace is $80\times{88}$ meters
We chose this workspace because the central part is relatively flat, while the two circled areas have interesting spatial variations.
We can vaguely see the environmental features in these circled spots from the satellite imagery.

\subsubsection{Hardware Setup}
We deploy the Autonomous Surface Vehicle~(ASV) shown in~\Cref{fig:field_system}.
The robot has a single-beam sonar pointing downward to collect depth measurements and a DJI Manifold 2-C computer for onboard computation.
The sonar is the Ping Sonar Altimeter and Echosounder from BlueRobotics.
Its maximum measurement distance is $50$ meters underwater, and the beam width is $30$ degrees.
It comes with a Python software interface, and we implemented its ROS driver, which is publicly available at \href{https://github.com/Weizhe-Chen/single_beam_sonar}{github.com/Weizhe-Chen/single\_beam\_sonar}.
The ASV from Clearpath Robotics has a built-in Extended Kalman Filter~(EKF) localization module that fuses the GPS signals and the UM6 Inertial Measurement Unit~(IMU) data.
The robot also has an embedded WiFi router for communication in the field.
The ASV is $1.3, 0.94$, and $0.34$ meters in length, width, and height, respectively, and is actuated by two thrusters at the rear.
It is a differential-drive robot, but its thrusters' maximum forward spinning speed is faster than the backward one.
We restrict the maximum linear velocity to $0.7$ meters per second and send linear and angular velocities to the robot to track an informative waypoint using a PD controller available at \href{https://github.com/Weizhe-Chen/tracking_pid}{github.com/Weizhe-Chen/tracking\_pid}.
Since the localization is unreliable, the robot only needs to reach a two-meter-radius circle centered at the waypoint.

\subsubsection{Results}
\Cref{fig:field_snapshot} shows the snapshots of the model prediction, uncertainty, and sampling path at different stages.
We can see that the prediction uncertainty is effectively reduced after sampling.
Most of the samples~(\textit{i.e.}, yellow dots) are collected in critical regions with drastic elevation variations.
Such a biased sampling pattern allows the robot to model the general trend of smooth regions with a small number of samples while capturing the characteristic environmental features at a fine granularity.

\section{Limitations and Future Work}\label{sec:discussion}
Although the AK has the same asymptotic computational complexity as the RBF kernel, its empirical runtime is slower than that of the RBF kernel.
Thus one important future work is to speed up the computation.
We leverage heuristics to train the non-stationary kernels in our experiments, which can be improved by a more principled training scheme in the future.
Using a stationary kernel in non-stationary environments is just one example of \emph{model misspecification}.
Investigating the influence of other types of model misspecification on RIG is interesting.
For example, the Gaussian likelihood assumes no sensing outliers, and the observational noise scale is the same everywhere.
Developing proper ways to handle sensing outliers and modeling \emph{heteroscedastic noise} can be important future work for RIG.
We only tried neural-network parameterization for the weighting function and the membership function.
Comparing different parameterization methods for the AK is also valuable.
Although we have only showcased the efficacy of AK in elevation mapping tasks, it has potential to benefit other applications such as 3D reconstruction, autonomous exploration and inspection, as well as search and rescue.
Exploring its utility in these domains would be interesting.
Additionally, while we focused on non-stationary kernels in the spatial domain, developing spatiotemporal kernels is crucial for RIG in dynamic environments.

\section{Conclusion}\label{sec:conclusion}
In this paper, we investigate the uncertainty quantification of probabilistic models, which is decisive for the performance of RIG but has received little attention.
We present a family of non-stationary kernels called the Attentive Kernel, which is simple, robust, and can extend any stationary kernel to a non-stationary one.
An extensive evaluation of elevation mapping tasks shows that AK provides better accuracy and uncertainty quantification than the two existing non-stationary kernels and the stationary RBF kernel.
The improved uncertainty quantification guides the informative planning algorithms to collect more valuable samples around the complex area, thus further reducing the prediction error.
A field experiment demonstrates that AK enables an ASV to collect more samples in important sampling locations and capture the salient environmental features.
The results indicate that misspecified probabilistic models significantly affect RIG performance, and GPR with AK provides a good choice for non-stationary environments.

\section{Acknowledgement}\label{sec:acknowledgment}
We acknowledge the support of NSF with grant numbers 1906694, 2006886, and 2047169.
We are also grateful for the computational resources provided by the Amazon AWS Machine Learning Research Award.
The constructive comments by the anonymous conference reviewers are greatly appreciated.
We thank Durgakant Pushp and Mahmoud Ali for their help in conducting the field experiment.

\balance
\bibliographystyle{SageH}
\interlinepenalty=10000
\bibliography{references.bib}
\end{document}